\documentclass[lettersize,journal]{IEEEtran}
\usepackage{amsmath,amsfonts}
\pdfoutput=1

\usepackage[ruled,linesnumbered]{algorithm2e}

\makeatletter
\newcommand{\removelatexerror}{\let\@latex@error\@gobble}
\makeatother

\usepackage{array}
\usepackage[caption=false,font=normalsize,labelfont=sf,textfont=sf]{subfig}
\usepackage{textcomp}
\usepackage{stfloats}
\usepackage{url}
\usepackage{hyperref}
\usepackage{verbatim}
\usepackage{graphicx}
\usepackage{subfloat}
\usepackage{tikz,xcolor}
\usepackage{cite}
\usepackage{multirow} 
\hypersetup{hidelinks,
	colorlinks=true,
	allcolors=black,
	pdfstartview=Fit,
	breaklinks=true}
\definecolor{lime}{HTML}{A6CE39}
\DeclareRobustCommand{\orcidicon}{
\begin{tikzpicture}
\draw[lime, fill=lime] (0,0)
circle[radius=0.16]
node[white]{{\fontfamily{qag}\selectfont \tiny \.{I}D}};
\end{tikzpicture}
\hspace{-2mm}
}
\foreach \x in {A, ..., Z}{%
\expandafter\xdef\csname orcid\x\endcsname{\noexpand\href{https://orcid.org/\csname orcidauthor\x\endcsname}{\noexpand\orcidicon}}
}

\begin{document}

\title{A Self-Distillation Embedded Supervised Affinity Attention Model for Few-Shot Segmentation}

\author{Qi~Zhao\hspace{-1.5mm}\orcidA{},~\IEEEmembership{Member,~IEEE,}
        Binghao~Liu\hspace{-1.5mm}\orcidB{}, 
        Shuchang~Lyu\hspace{-1.5mm}\orcidC{},~\IEEEmembership{Graduate Student Member,~IEEE,}
        \\and~Huojin~Chen\hspace{-1.5mm}\orcidD{}
\thanks{This work was supported in part by the National Natural Science Foundation of China under Grant 62072021. (Corresponding author: Huojin Chen)}
\thanks{Q. Zhao, B. Liu, S. Lyu are with the Department
of Electronic and Information Engineering, Beihang University, 37 Xueyuan Road, Haidian District, Beijing, P.R. China, 100191. (e-mail: zhaoqi@buaa.edu.cn, liubinghao@buaa.edu.cn, lyushuchang@buaa.edu.cn).}
\thanks{H. Chen is with the College of Art and Design, Beijing University of Technology, 100 Pingleyuan, Chaoyang District, Beijing, P.R. China, 100124. (e-mail: chenhuojin@bjut.edu.cn).}}

\markboth{Journal of \LaTeX\ Class Files,~Vol.~14, No.~8, August~2021}%
{Shell \MakeLowercase{\textit{et al.}}: A Sample Article Using IEEEtran.cls for IEEE Journals}


\maketitle

\begin{abstract}
Few-shot segmentation focuses on the generalization of models to segment unseen object with limited annotated samples. However, existing approaches still face two main challenges. First, huge feature distinction between support and query images causes knowledge transferring barrier, which harms the segmentation performance. Second, limited support prototypes cannot adequately represent features of support objects, hard to guide high-quality query segmentation. To deal with the above two issues, we propose self-distillation embedded supervised affinity attention model to improve the performance of few-shot segmentation task. Specifically, the self-distillation guided prototype module uses self-distillation to align the features of support and query. The supervised affinity attention module generates high-quality query attention map to provide sufficient object information. Extensive experiments prove that our model significantly improves the performance compared to existing methods. Comprehensive ablation experiments and visualization studies also show the significant effect of our method on few-shot segmentation task. On COCO-$20^i$ dataset, we achieve new state-of-the-art results. Training code and pretrained models are available at \href{https://github.com/cv516Buaa/SD-AANet}{https://github.com/cv516Buaa/SD-AANet}.
\end{abstract}


\section{Introduction}
\begin{figure}[htbp]
    \centering
    \includegraphics[width=1.0\linewidth]{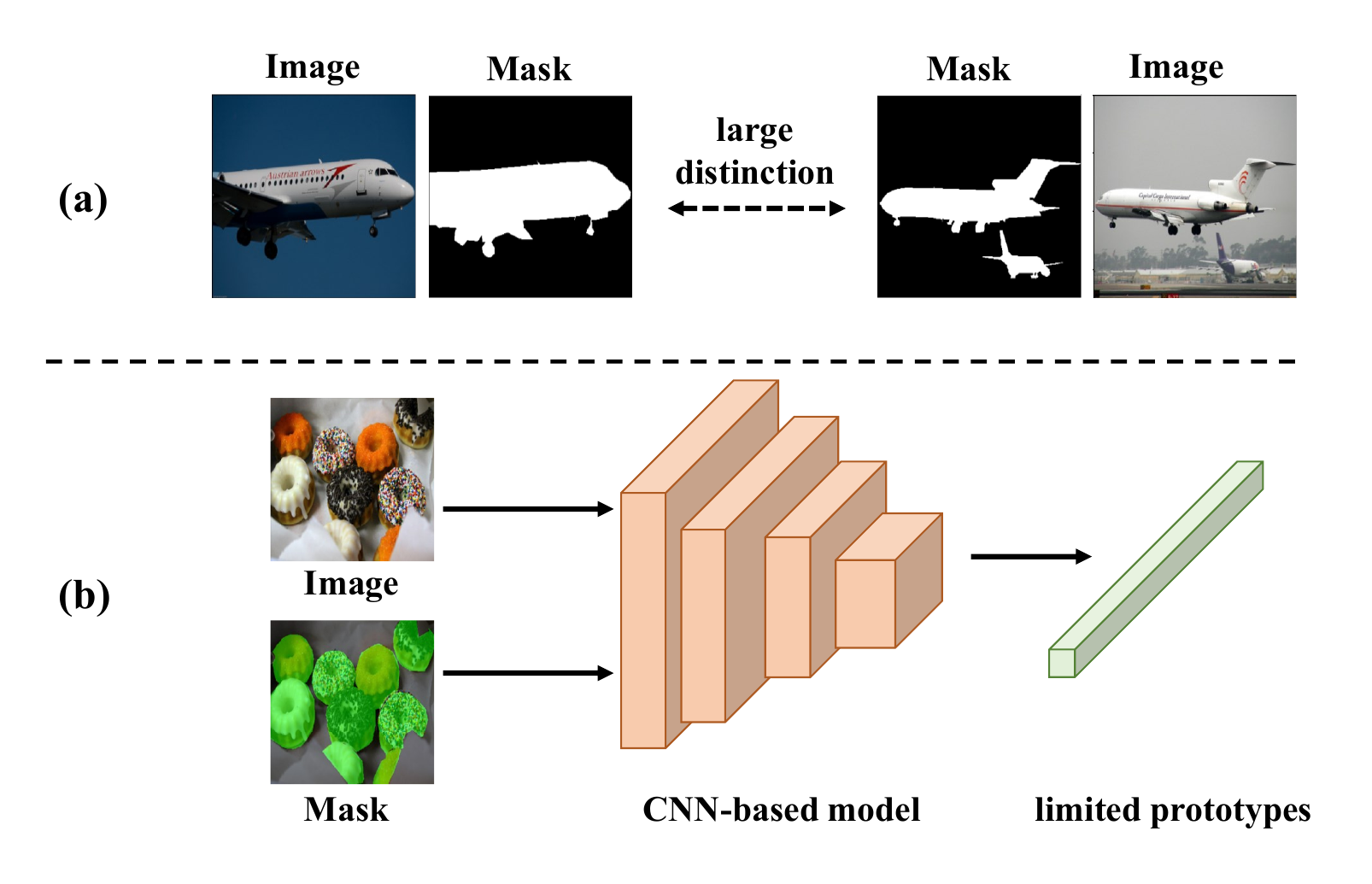}
    \caption{Two main challenges in few-shot segmentation task. (a) The huge feature distinction between support and query caused by differences on shooting angle, lighting conditions, shape and color of objects with same category. (b) Limited support prototypes cannot adequately represent features of support objects.}\label{Fig1}
\end{figure}

\IEEEPARstart{S}{emantic} segmentation, as a significant computer vision task, aims to assign a class label to each pixel in the image. Fully convolutional networks (FCNs) \cite{FCN} have been the pioneer to handle this task in an end-to-end manner, then various of deep neural network models \cite{SegNet, UNet, RefineNet, DeepLabV2, PSPNet, seg_tcds_1} have made great improvements recently. However, massive pixel-level annotated data required in semantic segmentation leads to expensive annotation cost. In addition, these methods have a dramatically performance decline when meeting unseen classes. In contrast, human cognitive ability can easily accurately complete complex tasks such as recognition based on the existing knowledge and a small amount of new labeled data.
\par To address the above-mentioned issues, few-shot segmentation \cite{OSLSM} is proposed, using limited annotated data to segment unseen classes. Different from fully supervised semantic segmentation, few-shot segmentation splits the whole data into support set and query set. The support set in this task provides meaningful and critical features of certain class to guide method extracting target with same class in query set.
\par Current few-shot segmentation methods are mainly based on metric learning, containing two main technical routes: affinity learning \cite{PGNet, BriNet, DAN} and prototypical learning \cite{SGOne, CANet, PMM}. Affinity learning acquires feature of support object with the help of support mask. Then each pixel-wise support feature is matched to query feature through various correlation measure operations, guiding query segmentation.
\par Prototypical learning methods usually use one or few prototypes to represent support object feature and guide segmentation of query target. These methods adopt masked global average pooling (masked GAP) \cite{SGOne} to obtain support prototype. Generally, the support prototype is combined with query feature through correlation metrics to realize high-quality segmentation. 
\par
\begin{figure}
    \centering
    \includegraphics[width=0.95\linewidth]{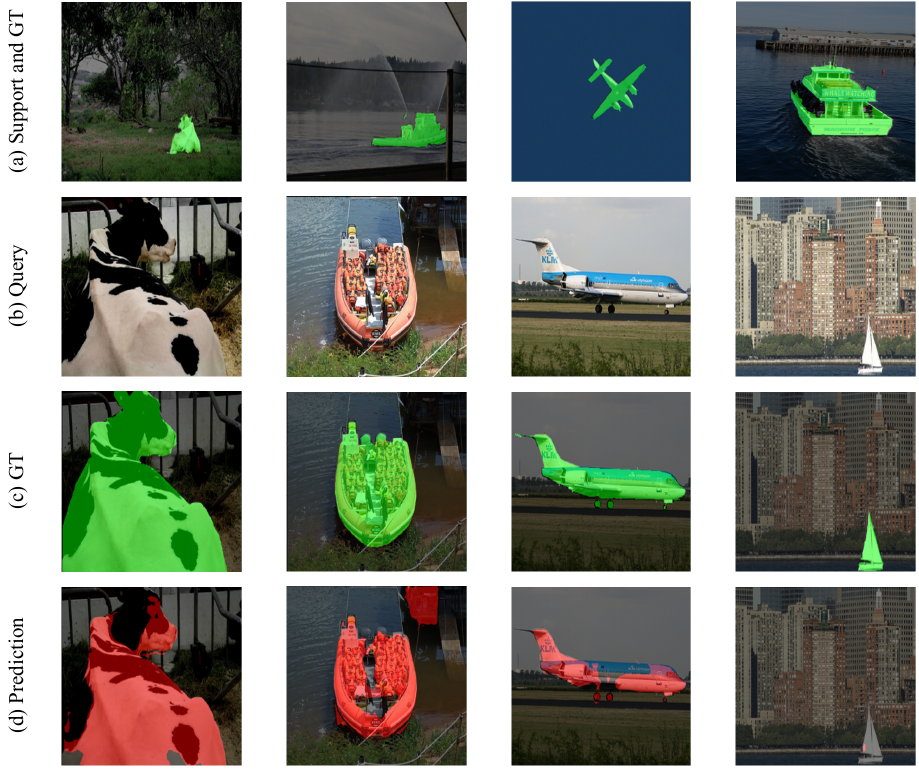}
    \caption{Some failure cases of current methods. From top to bottom: (a) support image and its ground truth, (b) query image, (c) ground truth of query image, (d) prediction of current methods.}\label{Fig2}
\end{figure}

 However, there are still two major challenges need to be solved in few-shot segmentation task. First, the appearance of objects in images may have different quantities, perspectives, illumination intensities, etc. The feature distinction between support and query dramatically reduces the segmentation performance. Second, limited support prototypes are incapable of providing sufficient representative information. Some typical failure cases of existing methods can be seen in Fig. \ref{Fig2}.

\par To deal with issue caused by the image distinctions, we attempt to creatively introduce the self distillation method into few-shot segmentation task. We propose a novel module named self-distillation guided prototype generating module (SDPM), which adopts self-distillation approach to bridge the gap between support and query features, finding commonalities between two features. SDPM takes support label, support feature and query feature as inputs, outputting a channel reweighting query feature and a support prototype with intrinsic class feature.
\par In few-shot segmentation task, one or few support prototypes can not provide sufficient representative information to segment the query target. So we design the supervised affinity attention module (SAAM), a CNN-based end-to-end module which can be simply embedded in deep CNN models and introduces negligible computation cost. SAAM has the same inputs as SDPM, and aims to generate an affinity attention map to give a prior prediction of query target.
\par Based on the two modules mentioned above, we propose the Self-Distillation embedded Affinity Attention network (SD-AANet) to produce intrinsic prototype and affinity attention map efficiently. Extensive experiments show that our SD-AANet achieves state-of-the-art performance on COCO-$20^i$ and  comparable state-of-the-art results on Pascal-$5^i$.
\par Our contributions are summarized as follows:
\begin{itemize}
\item We propose the SDPM to generate an intrinsic prototype by self-distillation approach, which can efficiently align the features of support and query. Otherwise, our SAAM helps to produce a query attention map to teach decoder where to focus. Through combining SDPM and SAAM, SD-AANet can better address the two challenges mentioned above.
\item SD-AANet achieves new state-of-the-art results on COCO-$20^i$ \cite{COCO, FWB} datasets (mIoU for 1-shot: 40.9\%, 5-shot: 48.7\%) and comparable state-of-the-art results on PASCAL-$5^i$ \cite{OSLSM} (mIoU for 1-shot: 62.9\%) for few-shot segmentation task.
\end{itemize}

\begin{figure*}[htbp]
    \centering
    \includegraphics[width=0.95\linewidth]{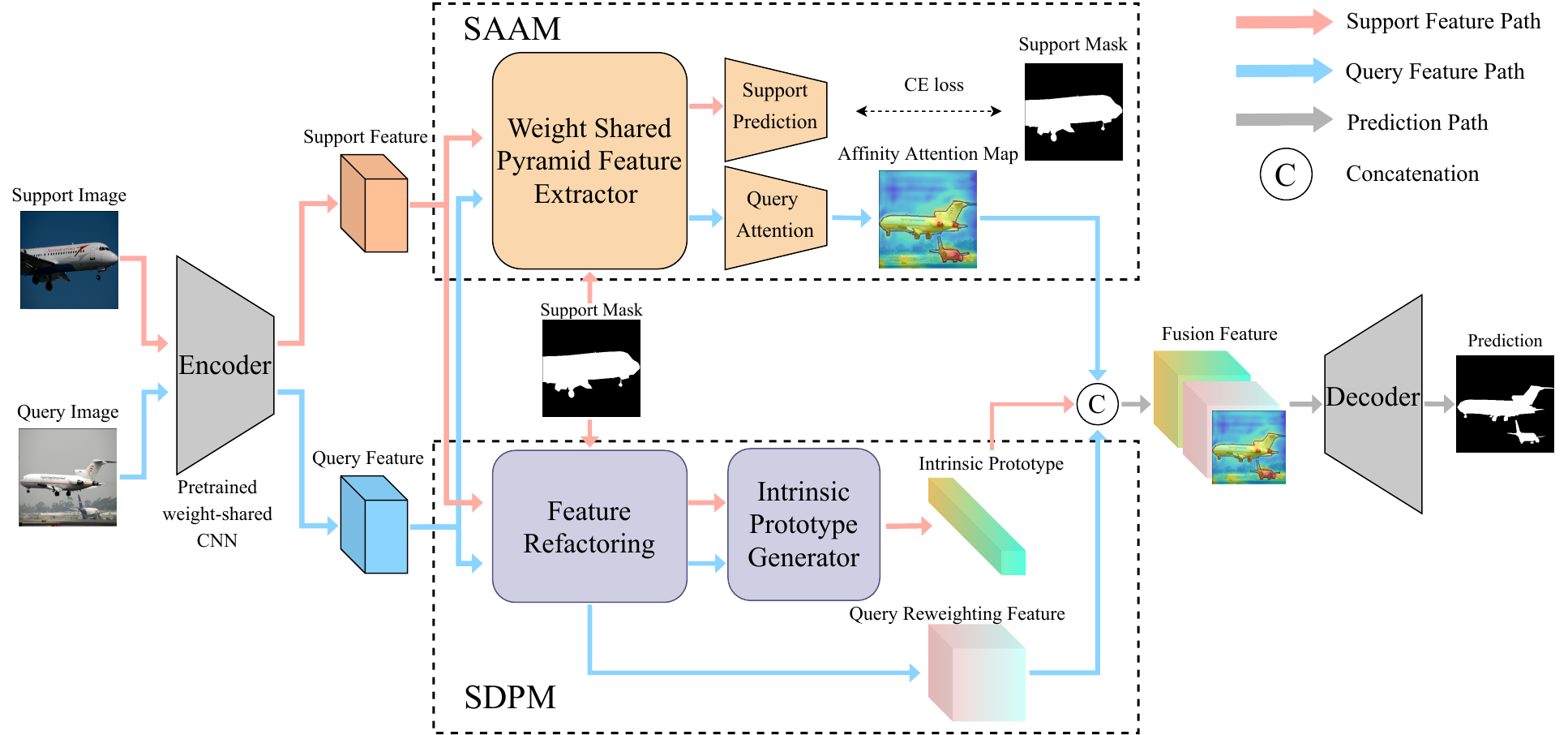}
    \caption{Architecture of SD-AANet. In SD-AANet, middle-level features of CNN backbone and support mask are input to SDPM and SAAM. SDPM uses support prototype to realize channel reweighting on support and query features. Then self-distillation approach in SDPM produces intrinsic support prototype. SAAM introduces support ground truth supervision to a learnable pyramid feature extractor, producing high-quality query attention map. Finally, the fusion of intrinsic prototype, query feature and query attention goes through the decoder to predict the final result.}\label{Fig3}
\end{figure*}

\section{Related Work}
\subsection{Semantic Segmentation}
Semantic Segmentation aims to predict a semantic category for each pixel in image. Convolutional Neural Network (CNN) based methods have made great progress in semantic segmentation field. Fully convolutional network (FCN) \cite{FCN} replaces fully connected layers with convolutional layers, achieving semantic segmentation in an end-to-end manner. SegNet \cite{SegNet} and UNet \cite{UNet} employ symmetric ``Encode-Decoder'' architectures to map the original image to the same-size predictions. PSPNet \cite{PSPNet} integrates pyramid pooling module into several baseline architectures like ResNet \cite{ResNet, Identity}) to obtain contextual information from different scales by using different kernel-sized pooling layers. Chen et al. \cite{6, 7} employ dilated convolution to expand the receptive field. In addition, some works focus on attention mechanism. PSANet \cite{PSANet} proposes a point-wise spatial attention to explore better connection information between pixels. DANet \cite{DANet} adopts position attention module and channel attention module to learn position and channel inter-dependencies. CCNet \cite{CCNet} adopts a criss-cross attention module to capture contextual information from full-image dependencies. However, well-performed semantic segmentation networks need a large amount of annotated data as training samples which are expensive to obtain.
\subsection{Few-shot learning}
Few-shot Learning seeks to recognize new objects with only few annotated samples. In this field, as an interpretable approach, metric learning \cite{Siamese, Matching, tcds_2} is widely used. Koch et al. \cite{Siamese} propose a siamese architecture which shows great performance on k-shot image classification tasks. This architecture can also be extended to deal with k-shot semantic segmentation. Meta learning \cite{Memory} enables machine to quickly acquire useful prior information from limited labeled samples. Meta-learning LSTM \cite{LSTM} and Model-Agnostic \cite{Model-Ag} methods apply recurrent neural network (RNN) to represent and store the prior information to handle the few-shot problem. To own the advantage of both two methods, ProtoMAML \cite{ProtoMAML} combines the complementary strengths of metric-learning and gradient-based meta-learning methods.
\subsection{Few-shot Segmentation}
Few-shot Semantic Segmentation aims at performing dense pixel-wise classification for unseen classes. Shaban et al. \cite{OSLSM} are the pioneers to officially define the few-shot semantic segmentation problem. They propose a two-branch architecture (OSLSM) to produce a binary mask for the new semantic class with dot-similarity manner. SG-One \cite{SGOne}, which is now a benchmark architecture in one-shot segmentation task, proposes an architecture that consists of a guidance branch and a segmentation branch. Based on two branchs design of SG-One \cite{SGOne},~\cite{PANet, FWB, PGNet, PPNet, CANet, PMM, SCL} further promote the few-shot segmentation performance. PFENet \cite{PFENet} proposes a training-free prior generation process to produce prior segmentation attention for the model, and a feature enrichment module to enrich query features with the support features. ASR \cite{ASR} reformulates few-shot segmentation as a semantic reconstruction problem and converts base class features into a series of basic vectors. HSNet \cite{HSNet} introduces 4D convolutions to extract diverse features from different levels of intermediate convolutional layers. ASNet \cite{ASNet} trains a learner to construct class-wise foreground maps for multi-label classification and pixel-wise segmentation. NTRENet \cite{NTRENet} explicitly mines and eliminates background and distracting objects regions for better segmentation. MSANet \cite{MSANet} exploits multiple feature-maps of support images and query images to estimate accurate semantic relationships. However, the feature distinction and the weak representation of limited support prototypes still hinder performance. Our SD-AANet utilizes self-distillation, prototypical learning and affinity learning, solving problems above and achieving performance improvements.

\section{Proposed Method}
In this section, we first briefly describe the definition of the few-shot segmentation task in Subsection \ref{3a}. Then in Subsection \ref{3b} and Subsection \ref{3c}, we introduce our self-distillation guided prototype generating module (SDPM) and supervised-based affinity attention module (SAAM) in details respectively. Finally, in Subsection \ref{3d}, we discuss optimization details and multi-class segmentation application of our proposed self-distillation embedded affinity attention model (SD-AANet).

\begin{figure*}[htbp]
    \centering
    \includegraphics[width=0.9\linewidth]{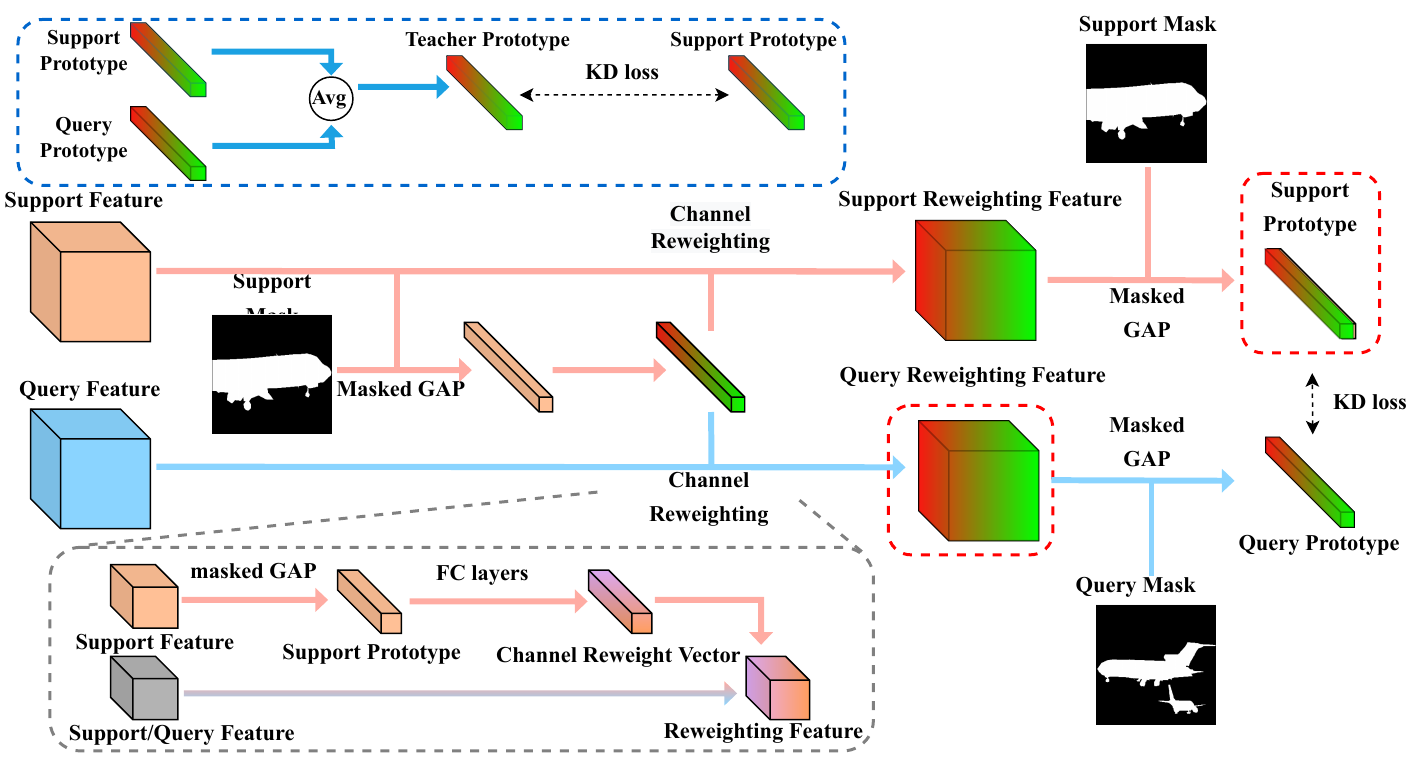}
    \caption{SDPM first applies masked GAP to generate support prototype, then it uses support prototype to produce channel reweighting vector. Channels of both support feature and query feature are reweighting by above-mentioned reweighting vector. After that, new support prototype and query prototype are generated by masked GAP, then self-distillation approach is used between two prototypes to produce intrinsic support prototype. In order to promote learning process of model, teacher vector in self-distillation approach is the average of support prototype and query prototype, as shown in blue dotted box. The ouputs of SDPM are query channel reweighting feature and support prototype shown in red dotted boxes.}\label{Fig4}
\end{figure*}

\subsection{Problem Setting}
\label{3a}
Few-shot segmentation is proposed to segment target of unseen classes under the guidance of limited annotated samples of the same classes. The dataset is split into two sets, training set $D_{train}$ and test set $D_{test}$, taking the class as the split standard. Defining the classes in $D_{train}$ as $C_{train}$ and the classes in $D_{test}$ as $C_{test}$, the two sets do not intersect, which means $C_{train}\bigcap C_{test}=\phi$.
\par Training and testing processes of few-shot segmentation can be seen as episodes. The episode paradigm was proposed in \cite{Episode}, and Shaban et al. \cite{OSLSM} first introduce it to few-shot segmentation. Each episode is consist of a support set $S$ and a query set $Q$ with the same class $C$. There are $K$ samples in the support set $S$, which is formulated as $S = \left \{ (I_{s}^{1}, M_{s}^{1}), (I_{s}^{2}, M_{s}^{2}), \cdots, (I_{s}^{K}, M_{s}^{K}) \right \}$. Each image-label pair ${(I_{s}^{i}, M_{s}^{i})}$ represents a sample in $S$, where $I_{s}^{i}$ and $M_{s}^{i}$ are the support image and its ground truth respectively. Similar to the support set $S$, query set $Q$ has one sample $(I_{q}, M_{q})$, where $I_{q}$ and $M_{q}$ are the query image and its ground truth respectively, having the same class $C$ with the support set $S$. The input of the model is a pair of query image $I_{q}$ and support set $S$, formulated as $\left \{ I_{q}, S \right \} = \left \{ I_{q}, (I_{s}^{1}, M_{s}^{1}), (I_{s}^{2}, M_{s}^{2}), \cdots, (I_{s}^{K}, M_{s}^{K}) \right \}$. Query ground truth $M_{q}$ is invisible in training stage and it is used to evaluate the performance of methods.

\subsection{Self-distillation Guided Prototype Generating Module}
\label{3b}
Current prototypical learning methods, such as PFENet \cite{PFENet}, approach a great performance on PASCAL-$5^i$ and COCO-$20^i$, outperforming previous works by a large margin. Prototypes generated by these methods can efficiently guide the segmentation of query target. However, there are large feature differences between support and query targets. So we need to align the features of support and query. Objects always have two types of features, intrinsic features which commonly exist in all objects of this class and unique features which may distinct in different objects. Take aeroplane as an example, all aeroplanes are made by metal and have wings. These features existing in all aeroplanes can be seen as intrinsic features. As the differences of shooting angle and lighting conditions, the shape and color of aeroplanes can be different, so they are unique features. Normally, humans have the cognitive ability to easily spot intrinsic features and apply them to subsequent tasks. For similar purposes, in few-shot segmentation, we need to find representative features of support and query images containing abundant intrinsic features.
\par The knowledge distillation approach proposed by Hinton et al. \cite{KD_1} greatly inspires us to transfer the knowledge between support and query prototypes. Zagoruyko et al. \cite{Attention_KD} and He et al. \cite{Affi_KD} expand the knowledge distillation technique by distilling attentions in middle layers. Fukuda et al. \cite{Multi_KD} propose integrating multiple teacher networks to teach the student network. Lyu et al. \cite{Lyu} realize the knowledge distillation in a single deep neural network, where student network is a part of the teacher network.
\par Inspired by the above methods, we introduce self-distillation approach to prototypical learning method, aiming to extract intrinsic features and aligning the features of support prototype and query prototype.
\subsubsection{Support Guided Channel Reweighting}
Different from the SENet \cite{SENet} which uses global feature to reweight channels of feature map, we alternatively adopt support prototype with better suitability for obtaining object-related information. As shown in Fig. \ref{Fig4}, architecture in gray dotted box is the variation of SE Module. Support image $I_{s}$ and query image $I_{q}$ with same class go through a shared backbone CNN, denoted as ${\mathcal {F}}\left ( \cdot  \right )$. $M_{s}$ and $M_q$ denote the ground truths of support and query, $F_{s}$ and $F_{q}$ denote the support feature and query feature which are outputs of middle-level layers of the backbone CNN:
\begin{equation}
  F_s = {\mathcal {F}}\left ( I_{s} \right ), F_q = {\mathcal {F}}\left ( I_{q} \right )
\label{eq1}
\end{equation}
note that $I_{s}$ and $I_{q}$ have same shape $n \times c \times h \times w$, in which $n$, $c$, $h$ and $w$ represent batch size, number of channels, height and width of the feature map.
\par Then support prototype is generated by masked GAP by calculating the average vector of the features in object area in feature map:
\begin{equation}
    p_s = \mathcal{F}_p \left ( F_s, M_s\right) = \frac{\sum_{i = 1}^{h}\sum_{j = 1}^{w}F_{s}^{\left ( i, j \right )}\left [ M_{s}^{\left ( i, j \right )}== 1 \right ]}{\sum_{i = 1}^{h}\sum_{j = 1}^{w}\left [ M_{s}^{\left ( i, j \right )}== 1 \right ]}
\label{eq2}
\end{equation}
where $i$ and $j$ denote the index of row and column, $F_{s}^{\left ( i, j \right )}$ denotes the position at row $i$, column $j$ in support feature map and $M_{s}^{\left ( i, j \right )}$ denotes the position at row $i$, column $j$ in support ground truth. $\mathcal{F}_p\left ( \cdot  \right )$ denotes the masked GAP operation. To guarantee the correctness of the Eq. \ref{eq2}, $M_s$ is resized to the same height and width with the support feature map. $\left [ \cdot  \right ]$ denotes Iverson bracket, a notation that signifies a number that is 1 if the condition in square brackets is satisfied, and 0 otherwise, i.e.
\par Acquired support prototype is then input to a series of fully connected layers (FC layers) to learn contributions of each feature channel, and there are ReLU functions between FC layers. Output of the FC layers is also a vector having the same number of channel with support feature and query feature. We have $v_s = {\mathcal FC \left ( p_s \right )}$, where ${\mathcal FC \left ( \cdot  \right )}$ and $v_s$ denote the FC layers and output channel reweighting vector respectively.
\par The channel reweighting vector $v_s$ scales channels of support feature and query feature, according to each channel's importance. Instead of using SE Block directly, we adopt a feature fusion strategy by using the average of scaled feature and input feature as the final channel reweighting feature:
\begin{equation}
    \tilde{F_s} = \frac{{\mathcal {F}}_{scale} \left ( v_s, F_s \right ) + F_s}{2},\tilde{F_q} = \frac{{\mathcal {F}}_{scale} \left ( v_s, F_q \right ) + F_q}{2}
\label{eq3}
\end{equation}
where ${\mathcal {F}}_{scale} \left ( v_s, F_s \right )$ denotes the channel scale function and $\tilde{F_s}$ denotes the final channel reweighting feature, so do the ${\mathcal {F}}_{scale} \left ( v_s, F_q \right )$ and $\tilde{F_q}$.


\subsubsection{Self-distillation embedded Method}
After Hinton et al. \cite{KD_1} first proposing the knowledge distillation in deep learning, many studies \cite{Lyu, Self-KD1, Self-KD2} have been conducted to let models learning from themselves. These approaches are named as self-distillation which aims to promote performance of model without external knowledge input.
\par Inspired by the above works, we introduce self-distillation approach to prototypical neural network, which can significantly improve few-shot segmentation performance by extracting intrinsic support feature.
\par To generate the intrinsic support prototype with the help of self-distillation approach, the average of support and query prototypes is used as the teacher. Masked GAP is employed to obtain both support prototype and query prototype from channel reweighting features as $p_{s}^{'} = \mathcal{F}_{p} \left ( \tilde{F_s}, M_s\right), p_{q}^{'} = \mathcal{F}_p \left ( \tilde{F_q}, M_q\right)$, where $p_{s}^{'}$ and $p_{q}^{'}$ denote the support prototype and query prototype after channel reweighting.
\par The query prototype and support prototype are adopted in self-distillation process. Both two prototypes can be seen as a combination of two parts feature, intrinsic feature and unique feature. So $p_{s}^{'}$ can be represented as $p_{s}^{'}\left( f_i, f_s\right)$, where $f_i$ and $f_s$ denote the intrinsic feature and support unique feature respectively. Similarly, query prototype $p_{q}^{'}$ can be represented as $p_{q}^{'}\left( f_i, f_q\right)$, and $f_q$ is the query unique feature. Following the knowledge distillation method, we apply the Kullback Leibler (KL) divergence loss to realize the supervision of support prototype:
\begin{equation}
    d_{s} = Softmax\left( p_{s}^{'}\right), d_{q} = Softmax\left( p_{q}^{'}\right)
\label{eq4}
\end{equation}

\begin{equation}
    L_{KD} = KL\left ( d_{t}\parallel d_{s} \right ) = \sum_{i = 1}^{c}d_{t}^{i}log\frac{d_{s}^{i}}{d_{t}^{i}}
\label{eq5}
\end{equation}
where $Softmax\left( \cdot\right)$ denotes the softmax function, $d_s$ and $d_q$ denote the outputs of the softmax function while inputs are $p_{s}^{'}$ and $p_{q}^{'}$. $d_{t}$ denotes the teacher prototype in knowledge distillation operation and it is equals to $d_{t} = \frac{d_{s} + d_{q}}{2}$. $L_{KD}$ in Eq. \ref{eq5} denotes the loss of self-distillation between support prototype and teacher prototype, and $KL\left( \cdot\right)$ denotes the KL divergence function.
\par Self-distillation approach enhances the consistency of support prototype and query prototype. Because the two prototypes are combinations of intrinsic feature and unique feature, the approach results in the lessen of unique feature and strengthen of intrinsic feature, which means $p_{s}^{'}\left( f_i, f_s\right)\rightarrow p_{s}^{'}\left( f_i\right)$. The $p_{s}^{'}\left( f_i\right)$ in formula denotes the support prototype with only intrinsic feature, more suitable for guiding the query segmentation.
\par SDPM can efficiently reduce the unique feature in support prototype and significantly ease the gap between support and query features. Extensive experiments show the improvement of performance more intuitively.

\subsubsection{K-shot Setting}
\par
\begin{figure}[tbp]
    \centering
    \subfloat[]{
		\label{Fig5.a}
		\includegraphics[width=0.9\linewidth]{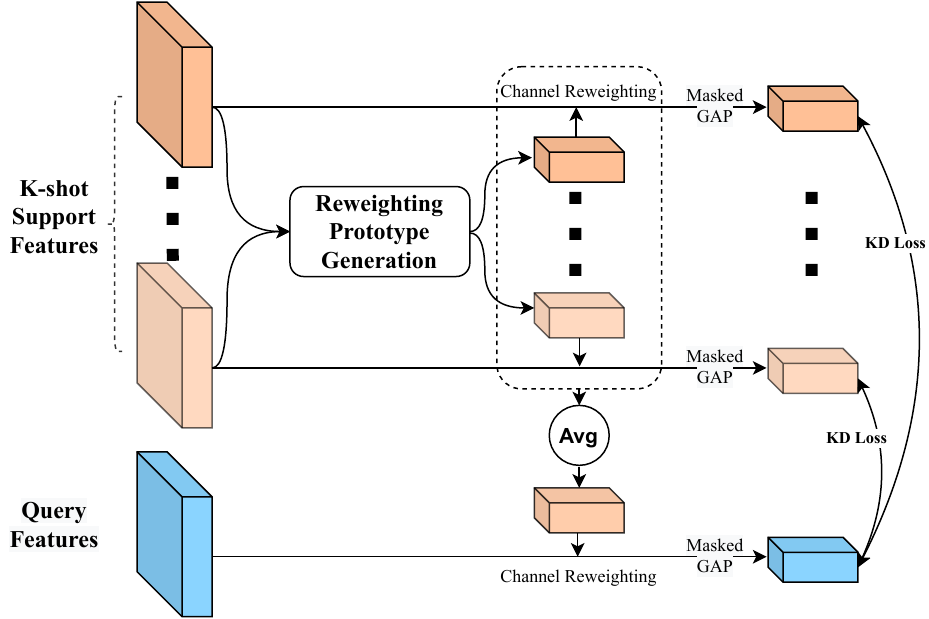}}
	\hspace{5mm}
	\subfloat[]{
		\label{Fig5.b}
		\includegraphics[width=0.9\linewidth]{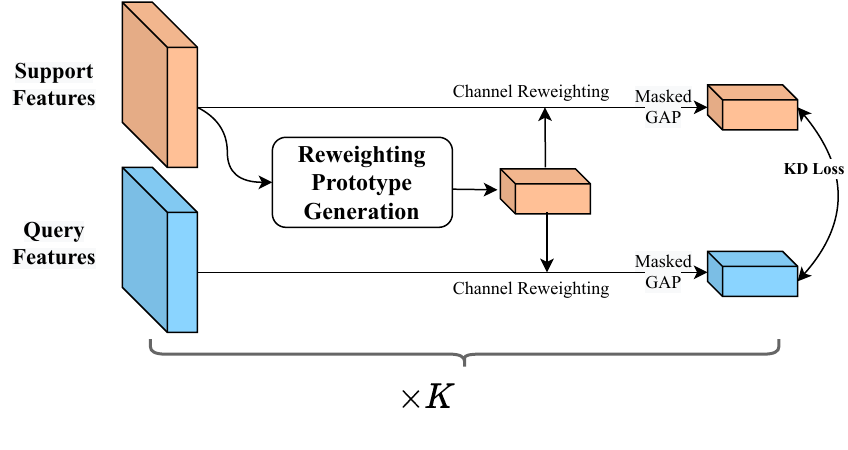}}
    \caption{Two strategies of SDPM in k-shot task, Integral Teacher Prototype Strategy (a) use the average of K reweighting vectors and masked GAP to produce the teacher prototype, while Separate Teacher Prototype Strategy (b) produces a exclusive teacher prototype for each support prototype.}
	\label{Fig5}
\end{figure}

 In addition to 1-shot segmentation, segmenting the query target under the guidance of K (greater than 1) support images is defined as K-shot segmentation. To extend SDPM to K-shot segmentation, this module needs to be modified appropriately. Because of the distinctions between K support images, teacher prototype extracted from query feature should supervise each support prototype separately. Depending on whether the teacher prototypes of K support prototypes are same, we design two strategies of SDPM for K-shot segmentation task, Integral Teacher Prototype Strategy and Separate Teacher Prototype Strategy. Details of two strategies are shown in Fig. \ref{Fig5}.

\par The core idea of Integral Teacher Prototype Strategy is applying the average of K reweighting vectors to scale each channel of query feature. Then masked GAP is adopted to extract teacher prototype, and K knowledge distillation losses are calculated between teacher prototype and each support prototype. The final self-distillation loss is the average of K losses.
\begin{equation}
    \tilde{F_q} = \frac{{\mathcal {F}}_{scale} \left ( \sum_{i = 1}^{K}v_{s}^{i}, F_q \right ) + F_q}{2}
\label{eq6}
\end{equation}
\begin{equation}
    L_{KD}^{I} = \frac{1}{K}\sum_{i = 1}^{K}KL\left ( p_{q}^{'}\parallel v_{s}^{i} \right )
\label{eq7}
\end{equation}
where $v_{s}^{i}$ denotes the prototype of $i$-th support sample, $p_{q}^{'}$ denotes the query prototype generated from $\tilde{F_q}$, and $L_{KD}^{I}$ denotes the knowledge distillation loss of Integral Teacher Prototype Strategy.
\par Different from Integral Teacher Prototype Strategy, as shown in Fig. \ref{Fig5.b}, Separate Teacher Prototype Strategy produces an exclusive teacher prototype for each support prototype, via applying each reweighting vector to scale the query feature separately.
\begin{equation}
    \tilde{F_q^i} = \frac{{\mathcal {F}}_{scale} \left ( v_{s}^{i}, F_q \right ) + F_q}{2}
\label{eq8}
\end{equation}
\begin{equation}
    L_{KD}^{S} = \frac{1}{K}\sum_{i = 1}^{K}KL\left ( p_{q}^{i}\parallel v_{s}^{i} \right )
\label{eq9}
\end{equation}
where $\tilde{F_q^i}$ and $p_{q}^{i}$ denote the query reweighting feature and teacher prototype produced by $i$-th support reweighting vector. $L_{KD}^{S}$ denotes the knowledge distillation loss of Separate Teacher Prototype Strategy. Final output query feature $\tilde{F_q}$ of SDPM is the average of K query reweighting features, formulated as $\tilde{F_q} = \frac{1}{K}\sum_{i = 1}^{K}\tilde{F_q^i}$.
\par No matter which strategy is applied, the final output support prototype is the average of K support prototypes. The ablation experiments in Section \ref{ablation} show the performances of two strategies. In the end, we use Separate Teacher Prototype Strategy in our overall model.
\begin{figure}[tbp]
    \centering
    \includegraphics[width=1.0\linewidth]{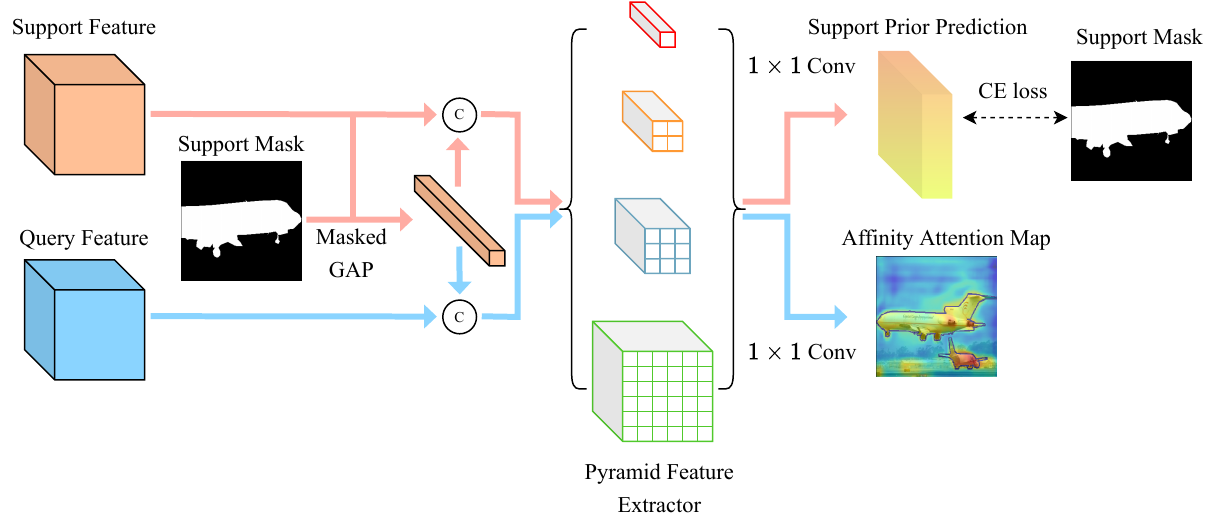}
    \caption{SAAM produces support prototype using masked GAP and concatenates support prototype to both support feature and query feature. PPM \cite{PSPNet} is adopted to extract features of two concatenated results. Then two $1\times 1$ convolutions with channel of 1 and 2 are applied to generate support prediction and query attention map. Support ground truth is used to supervise the support prediction, and the output of SAAM is query attention map.}
    \label{Fig6}
\end{figure}
\subsection{Supervised Affinity Attention Mechanism}
\label{3c}
\par Limited support prototypes can not provide enough target features, so we consider using the attention mechanism to provide more adequate prior information about the target for the task. Attention mechanism can effectively capture the location of object and let deep neural network models know where to focus, some previous works also utilize attention mechanism. PFENet \cite{PFENet} uses high-level features of both support and query to generate query attention map. By employing ImageNet \cite{ImageNet} pre-trained model as backbone and fixing its weights, the prior attention mechanism is training-free.
\par However, the ImageNet pre-trained model is produced to tackle the classification task, which is not suitable to generate attention map in few-shot segmentation task straightly. So we propose a supervised affinity attention mechanism (SAAM), which can solve the problem caused by unrepresentative of limited support features. The architecture of our SAAM is shown in Fig. \ref{Fig6}.
\subsubsection{Supervised Attention}
\par We first utilize masked GAP to obtain support prototype and expand it to the same spatial shape with support feature. Then the expanded prototype is concatenated to both support feature and query feature, we define the results as $F_{C,s}$ and $F_{C,q}$ respectively. Following, $F_{C,s}$ and $F_{C,q}$ are input to a pyramid feature extractor severally and outputs are defined as $F_{P, s}$ and $F_{P, q}$. We use Pyramid Pooling Module (PPM) \cite{PSPNet} as the pyramid feature extractor.
\par On the head of the PPM, there are two $1\times 1$ convolution layers (Convs) to generate support prediction and query attention map respectively. Support prediction is generated by $1\times 1$ Conv with two output channels. The $1\times 1$ Conv for query attention map generation has only one output channel. Then support label is applied to supervise the SAAM.
\begin{equation}
    L_{ce, s} = -\sum_{i = 1}^{h}\sum_{j = 1}^{w}\left ( M_{s}^{\left ( i, j \right )} \cdot log\left ( P_{s}^{\left ( i, j \right )} \right ) \right )
\label{eq10}
\end{equation}
where $L_{ce, s}$ is the cross entropy loss of support prediction in SAAM. $M_{s}^{\left ( i, j \right )}$ and $P_{s}^{\left ( i, j \right )}$ are $\left ( i, j \right )$ location of support mask and support prediction in SAAM.
\par The SAAM uses the whole support feature to guide the generation of query attention map, so the missing intrinsic features of support are learned under supervision. Combined with SDPM, the SD-AANet makes a trade-off during training and obtains richer intrinsic features to achieve better segmentation performance.

\subsubsection{K-shot Setting}
\par K-shot setting of SAAM is intuitive, because the only different part is support path. K support features go through SAAM severally, then each support prediction is supervised by its own label. Loss of K-shot SAAM is the average of K losses.
\begin{equation}
    L_{ce, s} = \sum_{i = 1}^{K}L_{ce, s}^i
\label{eq11}
\end{equation}
where $L_{ce, s}^i$ is the cross entropy loss of $i$-th support image.

\begin{table*}[tbp]
\caption{Comparison with state-of-the-art methods on PASCAL-$5^i$ with class mIoU metric. Baseline in table follows the PFENet \cite{PFENet}.}
\label{tab1}
\centering

\begin{tabular}{lcccccc|ccccc}
\\ \hline
\rule{0pt}{8pt} 
\multirow{2}{*}{Methods} & \multirow{2}{*}{Backbone} & \multicolumn{5}{c|}{1-shot}                                                   & \multicolumn{5}{c}{5-shot}                                                    \\ \cline{3-12} \rule{0pt}{8pt} 
                         &                           & Fold-0        & Fold-1        & Fold-2        & Fold-3        & Mean          & Fold-0        & Fold-1        & Fold-2        & Fold-3        & Mean          \\ \hline \rule{0pt}{8pt} 
OSLSM\cite{OSLSM}                    & VGG-16                    & 33.6          & 55.3          & 40.9          & 33.5          & 40.8          & 35.9          & 58.1          & 42.7          & 39.1          & 44.0          \\ \rule{0pt}{8pt}
co-FCN\cite{co-FCN}                   & VGG-16                    & 36.7          & 50.6          & 44.9          & 32.4          & 41.1          & 37.5          & 50.0          & 44.1          & 33.9          & 41.4          \\ \rule{0pt}{8pt}
SG-One\cite{SGOne}                   & VGG-16                    & 40.2          & 58.4          & 48.4          & 38.4          & 46.3          & 41.9          & 58.6          & 48.6          & 39.4          & 47.1          \\ \rule{0pt}{8pt}
AMP\cite{AMP}                      & VGG-16                    & 41.9          & 50.2          & 46.7          & 34.7          & 43.4          & 41.8          & 55.5          & 50.3          & 39.9          & 46.9          \\ \rule{0pt}{8pt}
PANet\cite{PANet}                    & VGG-16                    & 42.3          & 58.0          & 51.1          & 41.2          & 48.1          & 51.8          & 64.6          & 59.8          & 46.5          & 55.7          \\ \rule{0pt}{8pt}
PGNet\cite{PGNet}                    & ResNet50                  & 56.0          & 66.9          & 50.6          & 50.4          & 56.0          & 57.7          & 68.7          & 52.9          & 54.6          & 58.5          \\ \rule{0pt}{8pt}
FWB\cite{FWB}                      & ResNet101                 & 51.3          & 64.5          & 56.7          & 52.2          & 56.2          & 54.8          & 67.4          & 62.2          & 55.3          & 59.9          \\ \rule{0pt}{8pt}
CANet\cite{CANet}                    & ResNet50                  & 52.5          & 65.9          & 51.3          & 51.9          & 55.4          & 55.5          & 67.8          & 51.9          & 53.2          & 57.1          \\ \rule{0pt}{8pt}
CRNet\cite{CRNet}                    & ResNet50                  & -             & -             & -             & -             & 55.7          & -             & -             & -             & -             & 58.8          \\ \rule{0pt}{8pt}
RPMMs\cite{PMM}                    & ResNet50                  & 55.2          & 65.9          & 52.6          & 50.7          & 56.3          & 56.3          & 67.3          & 54.5          & 51.0          & 57.3          \\ \rule{0pt}{8pt}
SimPropNet\cite{SimPropNet}               & ResNet50                  & 54.8          & 67.3          & 54.5          & 52.0          & 57.2          & 57.2          & 68.5          & 58.4          & 56.1          & 60.0          \\ \rule{0pt}{8pt}
PPNet\cite{PPNet}                    & ResNet50                  & 47.8          & 58.8          & 53.8          & 45.6          & 51.5          & 58.4          & 67.8          & 64.9          & 56.7          & 62.0          \\ \rule{0pt}{8pt}
DAN\cite{DAN}                      & ResNet101                 & 54.7          & 68.6          & 57.8          & 51.6          & 58.2          & 57.9          & 69.0          & 60.1          & 54.9          & 60.5          \\ \rule{0pt}{8pt}
PFENet\cite{PFENet}                   & ResNet50                  & 61.7          & 69.5          & 55.4          & 56.3          & 60.8          & 63.1          & 70.7          & 55.8          & 57.9          & 61.9          \\ \rule{0pt}{8pt}
ASGNet\cite{ASGNet}                   & ResNet101                 & 59.8          & 67.4          & 55.6          & 54.4          & 59.3          & 64.6          & 71.3          & 64.2 & 57.3          & 64.4 \\ \rule{0pt}{6pt}
SCL\cite{SCL}                      & ResNet50                  & 63.0 & 70.0          & 56.5          & 57.7 & 61.8 & 64.5          & 70.9          & 57.3          & 58.7          & 62.9          \\ \rule{0pt}{6pt}
MMNet\cite{MMNet}                      & ResNet50 v2                 & 62.7 & 70.2          & 57.3          & 57.0 & 61.8 & 62.2          & 71.5          & 57.5          & 62.4          & 63.4          \\ \rule{0pt}{6pt}
CWT\cite{CWT}                      & ResNet50                  & 56.3 & 62.0          & 59.9          & 47.2 & 56.4 & 61.3          & 68.5          & 68.5          & 56.6          & 63.7          \\ \rule{0pt}{6pt}
CMN\cite{CMN}                      & ResNet50                  & 64.3 & 70.0          & 57.4          & 59.4 & 62.8 & 65.8          & 70.4          & 57.6          & 60.8          & 63.7          \\ \rule{0pt}{6pt}
ASR\cite{ASR}                      & ResNet50                 & 55.2 & 70.4          & 53.4          & 53.7 & 58.2 & 59.4          & 71.9          & 56.9          & 55.7          & 61.0  
\\ \rule{0pt}{6pt}
HSNet\cite{HSNet}                      & ResNet50                  & 64.3 & 70.7          & 60.3          & 60.5 & 64.0 & 70.3          & 73.2          & \textbf{67.4}          & \textbf{67.1}          & 69.5 
\\ \rule{0pt}{6pt}
ASNet\cite{ASNet}                      & ResNet50                  & \textbf{68.9} & 71.7          & \textbf{61.1}          & \textbf{62.7} & \textbf{66.1} & \textbf{72.6}          & \textbf{74.3}          & 65.3          & \textbf{67.1}          & \textbf{70.8}  
\\ \rule{0pt}{6pt}
NTRENet\cite{NTRENet}                      & ResNet50                  & \textbf{68.9} & 71.7          & 59.4          & 59.8 & 64.2 & 66.2          & 72.8          & 61.7          & 62.2          & 65.7  

\\ \hline \rule{0pt}{6pt} 
Baseline(ours)           & ResNet50                  & 59.5          & 70.0          & 55.9          & 56.1          & 60.4          & 63.7          & 70.5          & 57.2          & 57.5          & 62.2          \\ \rule{0pt}{6pt}
SD-AANet(ours)           & ResNet50                  & 62.7          & \textbf{71.8} & 58.8 & 58.1          & 62.9 & 65.5 & 71.6 & 62.5          & 62.3 & 65.5 \\ \hline
\end{tabular}
\end{table*}

\subsection{Optimization}
\label{3d}
Based on the SDPM and SAAM, we propose the Self-Distillation Embedded Supervised Affinity Attention Network (SD-AANet), as shown in Fig.\ref{Fig3}. For the whole model, we choose cross entropy loss $L_{ce}$ for the final segmentation prediction. Counting losses of SDPM and SAAM in, the total loss of SD-AANet is the combination of $L_{ce}$, $L_{KD}$ and $L_{ce, s}$ as
\begin{equation}
    L = L_{ce} + \alpha \cdot L_{KD} + \beta \cdot L_{ce, s}
\label{eq12}
\end{equation}
where $\alpha$ and $\beta$ are coefficients of $L_{KD}$ and $L_{ce, s}$, used to balance three loss compositions.

\subsection{Multi-class Few-shot Segmentation}

\par To further explore the potential of SD-AANet, we propose a new pipeline for segmenting multi-class objects simultaneously under few-shot setting. Because the fore-mentioned method and pipeline can not applied to segmenting multi-class objects straightly, we modify the decoder and design new training and testing pipeline. We describe the modification and the new pipeline under 1-shot setting as follow.
\par The input support images and masks are come from five different classes, at least one of which has same class with objects in query image. Then after going through encoder, SDPM and SAAM, there will be one query feature, five support prototypes and five attention maps. Before decoder, we concatenate these prototypes to one vector and add a MLP to reduce its dimension. Then we concatenate the obtained vector, query feature and five attention map as input of decoder. To make decoder segment objects from five classes simultaneously, we change the final output channel from 2 to 5. The output prediction has five channels, whose order is same as the order of concatenation of five support prototypes. The whole learning process of multi-class 1-shot segmentation can be seen in Algorithm. \ref{alg1}.

\begin{figure}[h]

 \removelatexerror
\begin{algorithm}[H]
\label{alg1}
\caption{The learning process of Multi-class 1-shot Segmentation}
  \KwIn{support set $S = \left \{ (I_{s}^{1}, M_{s}^{1}), \cdots, (I_{s}^{5}, M_{s}^{5}) \right \}$, query image $I_{q}$ and mask $M_q$}
  \KwOut{learnable parameters $p$ of SD-AANet}
  \For{$each\ batch\ i=0\ to\ total\ batch$}{
    Obtain support prototypes $v_{s}^{i}$ ($\ i=1,\dots, 5$); \\
    Get attention maps $m^{i}$ ($\ i=1,\dots, 5$) and query feature $F_q$; \\
    Concatenate $v_{s}^{i}$ to $v_{c}$; \\
    $v_{new} = MLP\left ( v_{c} \right )$; \\
    Concatenate $v_{new}$, $F_q$ and $m^{i}$ as $F_{new}$; \\
    $p = Decoder\left ( F_{new} \right )$; \\
    Calculate $L_{ce}$, $L_{KD}$ and $L_{ce,s}$; \\
    Train model with $L = L_{ce} + \alpha \cdot L_{KD} + \beta \cdot L_{ce, s}$; 
  }
  
\end{algorithm}
\end{figure}

\begin{table*}[htbp]
\caption{Comparison with state-of-the-art methods on COCO-$20^i$ with class mIoU metric. Baseline in table follows the PFENet \cite{PFENet}. Models with † are evaluated on the labels resized to $473\times 473$ size. Models without † are tested on labels with the original sizes.}
\label{tab2}
\centering
\begin{tabular}{lcccccc|ccccc}
\hline \rule{0pt}{8pt}
\multirow{2}{*}{Methods} & \multirow{2}{*}{Bcakbone} & \multicolumn{5}{c|}{1-shot}                                                   & \multicolumn{5}{c}{5-shot}                                                    \\ \cline{3-12}  \rule{0pt}{8pt}
                         &                           & Fold-0        & Fold-1        & Fold-2        & Fold-3        & Mean          & Fold-0        & Fold-1        & Fold-2        & Fold-3        & Mean          \\ \hline \rule{0pt}{8pt}
FWB\cite{FWB}                      & ResNet101                 & 19.9          & 18.0          & 21.0          & 28.9          & 21.2          & 19.1          & 21.5          & 23.9          & 30.1          & 23.7          \\ \rule{0pt}{8pt}
PANet\cite{PANet}                    & VGG-16                    & -             & -             & -             & -             & 20.9          & -             & -             & -             & -             & 29.7          \\ \rule{0pt}{8pt}
DAN\cite{DAN}                      & ResNet101                 & -             & -             & -             & -             & 24.4          & -             & -             & -             & -             & 29.6          \\ \rule{0pt}{8pt}
RPMMs\cite{PMM}                    & ResNet50                  & -             & -             & -             & -             & 30.6          & -             & -             & -             & -             & 35.5          \\ \rule{0pt}{8pt}
PPNet\cite{PPNet}                    & ResNet50                  & 28.1          & 30.8          & 29.5          & 27.7          & 29.0          & 39.0          & 40.8          & 37.1          & 37.3          & 38.5          \\ \rule{0pt}{8pt}
PFENet\cite{PFENet}                   & ResNet101                 & 34.3          & 33.0          & 32.3          & 30.1          & 32.4          & 38.5          & 38.6          & 38.2          & 34.3          & 37.4          \\ \rule{0pt}{8pt}
PFENet†\cite{PFENet}                  & ResNet101                 & 36.8          & 41.8          & 38.7          & 36.7          & 38.5          & 40.4          & 46.8          & 43.2          & 40.5          & 42.7          \\ \rule{0pt}{8pt}
ASGNet\cite{ASGNet}                   & ResNet50                  & -             & -             & -             & -             & 34.6          & -             & -             & -             & -             & 42.5          \\ \rule{0pt}{8pt}
SCL\cite{SCL}                      & ResNet101                 & 36.4          & 38.6          & 37.5          & 35.4          & 37.0          & 38.9          & 40.5          & 41.5          & 38.7          & 39.9          \\ \rule{0pt}{8pt}
MMNet\cite{MMNet}                   & ResNet50 v2                 & 34.9             & 41.0             & 37.2             & 37.0             & 37.5          & 37.0             & 40.3             & 39.3             & 36.0             & 38.2          \\ \rule{0pt}{8pt}
CWT\cite{CWT}                   & ResNet101                 & 30.3             & 36.6             & 30.5             & 32.2             & 32.4          & 38.5             & 46.7             & 39.4            & 43.2             & 42.0          \\ \rule{0pt}{8pt}
CMN\cite{CMN}                      & ResNet50                 & 37.9          & \textbf{44.8}          & 38.7          & 35.6          & 39.3          & 42.0          & 50.5          & 41.0          & 38.9          & 43.1          \\ \hline \rule{0pt}{8pt}
Baseline(ours)           & ResNet101                 & 35.5          & 41.5          & 37.6          & 36.1          & 37.7          & 41.5          & 47.3          & 45.2          & 41.6          & 43.9          \\ \rule{0pt}{8pt}
Baseline(ours)†          & ResNet101                 & 36.7          & 41.9          & 38.1          & 37.2          & 38.5          & 41.7          & 48.6          & 46.8          & 42.9          & 45.0          \\ \rule{0pt}{8pt}
SD-AANet(ours)           & ResNet101                 & 39.3 & 43.9          & 39.8          & 39.5          & 40.6          & 43.1 & 51.4          & 52.7          & 46.3          & 48.4          \\ \rule{0pt}{8pt}
SD-AANet(ours)†          & ResNet101                 & \textbf{39.6}          & 44.3 & \textbf{39.9} & \textbf{39.9} & \textbf{40.9} & \textbf{43.2}          & \textbf{51.9} & \textbf{52.9} & \textbf{46.6} & \textbf{48.7} \\ \hline \rule{0pt}{8pt}
\end{tabular}
\end{table*}

\section{Experiments}
\subsection{Implementation Details}
\subsubsection{Datasets}
The PASCAL-$5^i$ \cite{OSLSM} and COCO-$20^i$ \cite{COCO, FWB} datasets are used in experiments to evaluate our method. PASCAL-$5^i$ consists of two parts, PASCAL VOC 2012 \cite{PASCAL} and extended annotations from SDS datasets \cite{SDS}. There are 20 classes in original PASCAL VOC 2012 and SDS, and they are evenly divided into 4 folds, defined as Fold-$i$, $i \in \left \{ 0, 1, 2, 3 \right \}$. Each fold contains 5 classes following settings in OSLSM \cite{OSLSM}, and 1000 pairs of support-query are used in our test.
\par Following \cite{FWB}, COCO dataset, owning 80 classes totally, is also splited into 4 folds with 20 classes in each fold. The set of class indexes contained in fold-$i$ is written as $\left \{ 4x - 3 + i\right \}$, $i \in \left \{ 1, 2, \cdots, 20 \right \}$. Due to the number of images in COCO validation is 40137, which is much more than the PASCAL-$5^i$. So we randomly choose 4000 support-query pairs each fold during testing following \cite{FWB}, which can provide more reliable and stable results for 20 classes than 1000 pairs. 
\par To realize the few-shot segmentation, we use three folds 
to train and test the model on last fold for cross-validation. We alternatively choose different folds in testing to evaluate performance of our model, and we carry out five rounds of experiment and take the average to get the final experimental results.
\subsubsection{Experimental Setting}
We use PyTorch to construct our framework, and we apply ResNet50 and ResNet101 \cite{ResNet} as our backbones for PASCAL-$5^i$ and COCO-$20^i$ respectively. We choose the ResNet with atrous convolutions as the same with previous works \cite{FWB, CANet, A-MCG}. The ImageNet \cite{ImageNet} pretrained weights provided by PyTorch are used to initialize backbone networks. We use SGD as our optimizer. We set the momentum and weight decay to 0.9 and  0.0001 respectively. The 'poly' policy \cite{DeepLabV2} is adopted in our experiments to decay the learning rate by multiplying $\left ( 1 - \frac{current_{iter}}{max_{iter}} \right )^{power}$ where $power$ is set to 0.9. $\alpha$ and $\beta$ are set to 50 and 0.5 respectively. We use PFENet \cite{PFENet} as our baseline.
\par The experiments on PASCAL-$5^i$ train models for 200 epochs as \cite{PFENet}, while the initial learning rate and batch size are set to 0.0025 and 4. Because there are more images in training set of COCO-$20^i$, we train models for 50 epochs with 0.0005 and 8 for the initial learning rate and batch size respectively. We fix the parameters of backbone networks and update other parameters during training. Each example is processed with mirror operation and random rotation from -10 to 10 degrees. Finally, limited by equipment performance, we randomly crop $321\times 321$ patches from the processed images as training samples, which significantly reduces storage consumption and runtime. During evaluation, we resize the processed images to $321\times 321$ and pad zero to maintain the original aspect ratio of images. Then the prediction is resized back to original label sizes to evaluate performance. Following \cite{PFENet}, for COCO-$20^i$, we also resize the prediction to $473\times 473$ with respect to its original aspect ratio to make another evaluation. The single-scale results are output without multi-scale testing and any other post-processing. Our experiments are conducted on a NVIDIA GeForce RTX3090 GPU and Intel Xeon CPU 10900K.
\subsubsection{Evaluation Metrics}
Following \cite{FWB, CANet, PFENet}, we adopt class mean intersection over union (mIoU) as our evaluation metric, because the class mIoU is more reasonable than the foreground-background IoU (FB-IoU) \cite{CANet}. The formulation of class mIoU is $\frac{1}{C}\sum_{i=1}^{C}IoU_{i}$, where $C$ is the number of classes belong to each fold. So $C = 5$ for PASCAL-$5^i$ and $C = 20$ for COCO-$20^i$. The $IoU_{i}$ is intersection over union of $i$-th class.

\begin{table}[tbp]

\caption{The complexity and computational efficiency of Baseline and SD-AANet.}
\label{tab3}
\centering
\begin{tabular}{l|ccc}
\hline
\multicolumn{1}{c|}{Methods} & GPU memory & FPS   & \# learnable params \\ \hline
Baseline                     & 2216M      & 18.75 & 10.8M               \\
SD-AANet                     & 2483M      & 17.65 & 14.1M               \\ \hline
\end{tabular}
\end{table}

\subsection{Results}
As shown in Tables \ref{tab1} and \ref{tab2}, we adopt ResNet50 and ResNet101 to build our models for PASCAL-$5^i$ and COCO-$20^i$ respectively. And we report the class mIoU results to prove the performance of our proposed models. By incorporating the SDPM and SAAM, with the $321\times 321$ size of input images which is smaller than $473\times 473$ used in previous works, our SD-AANet still achieves comparable state-of-the-art results on PASCAL-$5^i$ and reaches new state-of-the-art results on COCO-$20^i$ for class mIoU metric.
\par In Table \ref{tab1}, we compare our model with other state-of-the-art methods on PASCAL-$5^i$. On this dataset, ASNet \cite{ASNet} and HSNet \cite{HSNet} achieve better results which are significantly higher than other methods. However, HSNet introduces 4D convolutions to integrate multi-level features, although center-pivot are used to decrease the space and time complexities, the costs are still huge. ASNet computes the correlations between each point in support feature and those in query feature, which can introduce non-negligible computational cost. Otherwise, our SD-AANet still shows some advantages in some categories, such as the result on Fold-1 for 1-shot task.

\begin{table}[tbp]
\caption{Ablation study of our proposed SDPM and SAAM on PASCAL-$5^i$ for 1-shot segmentation. ``SD-AANet'' means the model with both SDPM and SAAM.}
\label{tab4}
\centering
\begin{tabular}{l|cccc|c}
\hline \rule{0pt}{8pt}
Methods                           & Fold-0 & Fold-1 & Fold-2 & Fold-3 & Mean \\ \hline \rule{0pt}{8pt}
Baseline                          & 59.5   & 70.0   & 55.9   & 56.1   & 60.4 \\ \rule{0pt}{8pt}
Baseline + SAAM                   & 60.5   & 70.8   & 57.4   & 57.6   & 61.6 \\ \rule{0pt}{8pt}
Baseline + SDPM                   & 61.1   & 71.0   & 58.0   & 57.8   & 62.0 \\ \rule{0pt}{8pt}
SD-AANet & \textbf{62.7}   & \textbf{71.8}   & \textbf{58.8}   & \textbf{58.1}   & \textbf{62.9} \\ \hline 
\end{tabular}
\end{table}

\begin{table}[tbp]
\caption{Ablation study of using single-scale inference or multi-scale inference on PASCAL-$5^i$ for 1-shot segmentation.}
\label{tab5}
\centering
\begin{tabular}{l|cccc|c}
\hline \rule{0pt}{8pt}
 Methods      & Fold-0        & Fold-1        & Fold-2        & Fold-3        & Mean          \\ \hline \rule{0pt}{8pt}
        Single-scale & \textbf{62.7} & 71.8          & 58.8          & 58.1          & 62.9          \\ \rule{0pt}{8pt}
        Multi-scale  & 62.5          & \textbf{72.2} & \textbf{59.3} & \textbf{58.9} & \textbf{63.2} \\ \hline
\end{tabular}
\end{table}

\par In Table \ref{tab2}, our SD-AANet achieves new state-of-the-art performance on COCO-20$i$ for both 1-shot task and 5-shot task, and surpasses CMN \cite{CMN} by 1.6$\%$ and 5.6$\%$. Besides, the SDPM and SAAM improve the performance by 2.4$\%$ and 3.7$\%$ than our baseline.

\par We analyze the complexity and computational efficiency of the baseline model and SD-AANet in Tab. \ref{tab3}. Compared to the baseline model, our SD-AANet increases the GPU memory cost during inference phase and the number of learnable parameters by 12$\%$ and 30$\%$, respectively. Otherwise, SD-AANet slightly reduces the inference speed from 18.75 to 17.65 on Frames Per Second (FPS).

\par From what has been discussed above, due to the introduction of SDPM and SAAM, SD-AANet achieves superior results on few-shot segmentation task.

\subsection{Ablation Study}
\label{ablation}
\subsubsection{Ablation Study of SDPM and SAAM}
\label{Ablation One}
To quantitatively analyze the influence of SDPM and SAAM, we conduct an experiment about the performance of model w/ and w/o the SDPM and SAAM. Table \ref{tab4} shows the class mIoU results of each model on PASCAL-$5^i$ for 1-shot task.
\par It can be seen in Table \ref{tab4} that, compared to baseline, using only SAAM or SDPM improve the performance with class mIoU increases of 1.2$\%$ and 1.6$\%$ respectively. Adopting both SAAM and SDPM can further improve the performance, with 2.5$\%$ class mIoU gain.

\begin{figure*}[tbp]
    \centering
    \includegraphics[width=0.9\linewidth]{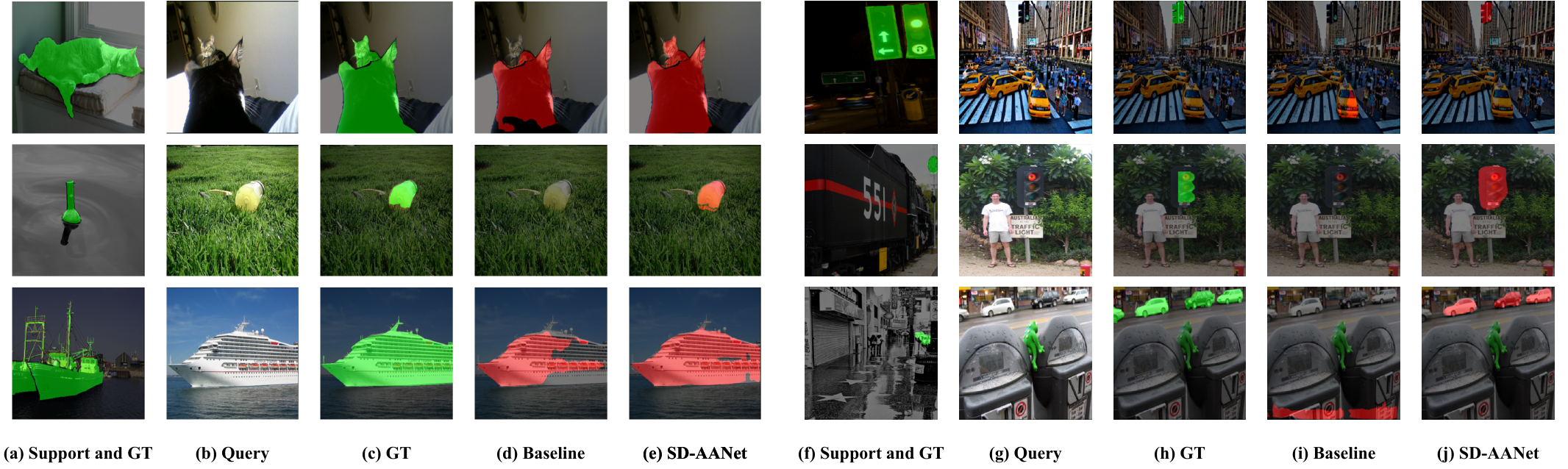}
    \caption{Qualitative results of the proposed SD-AANet and the baseline for 1-shot task. The columns from (a) to (e) is results on PASCAL-$5^i$, and the columns from (f) to (j) is results on COCO-$20^i$.}
    \label{Fig7}
\end{figure*}

\subsubsection{Ablation Study of Multi-scale Inference}
Table \ref{tab5} shows a comparison experiment between single-scale inference and multi-scale inference of SD-AANet. The experiment is conducted on PASCAL-$5^i$ for 1-shot task. In this experiment, we resize the logits before classify layer to $321\times 321$ and $473\times 473$, and adopt softmax operation to get two prediction with different size. Then we resize the $321\times 321$ prediction to $473\times 473$ by using bilinear interpolation. Finally the two $473\times 473$ predictions are added to get final prediction. The result shows that multi-scale inference can improve the performance slightly with 0.3$\%$ class mIoU gain. 

\begin{figure*}[htbp]
    \centering
    \includegraphics[width=0.9\linewidth]{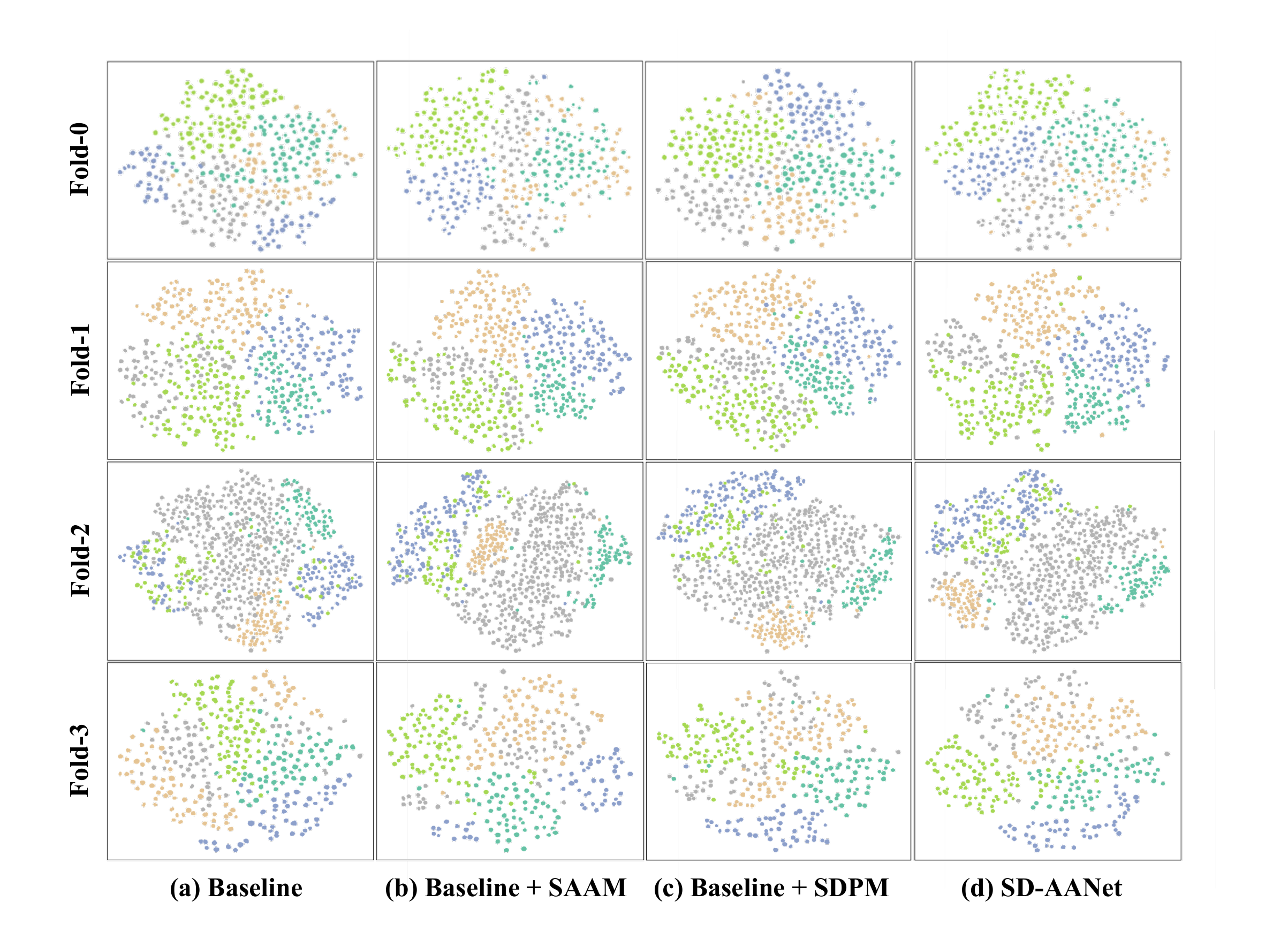}
    \caption{Visualization comparison study of support prototypes between t-SNE results. Each figure contains 5000 support prototypes generated from the same 5000 image pairs.}
    \label{Fig8}
\end{figure*}

\subsubsection{Ablation Study of 5-shot Strategy}
Table \ref{tab6} studies the influence of different 5-shot strategy of SDPM. We design two strategies for SDPM in 5-shot task, Integral Teacher Prototype Strategy and Separate Teacher Prototype Strategy, named Integral Strategy and Separate Strategy later. Comparison experiment shown in Table \ref{tab6} indicates the Separate Strategy achieves more outstanding result than Integral Strategy. It means that assign a unique teacher prototype to each support prototype can facilitate the production of intrinsic support prototype, which promotes the segmentation performance of query target.

\begin{table}[tbp]
\caption{Ablation study of SDPM with Integral Teacher Prototype Strategy of Separate Teacher Prototype Strategy on PASCAL-$5^i$ for 5-shot segmentation.}
\label{tab6}
\centering
\begin{tabular}{l|cccc|c}
\hline \rule{0pt}{8pt}
Methods           & Fold-0        & Fold-1        & Fold-2        & Fold-3        & Mean          \\ \hline \rule{0pt}{8pt}
        Integral Strategy & 64.9          & 70.8          & 61.7 & 61.1          & 64.6          \\ \rule{0pt}{8pt}
        Separate Strategy & \textbf{65.5} & \textbf{71.6} & \textbf{62.5}          & \textbf{62.3} & \textbf{65.5} \\ \hline
\end{tabular}
\end{table}

\begin{table}[tbp]
\caption{Ablation study of baseline model and SD-AANet for mutli-class 1-shot segmentation.}
\label{tab7}
\centering
\begin{tabular}{c|cccc|c}
\hline
Methods  & Fold-0                      & Fold-1                      & Fold-2 & Fold-3 & Mean \\ \hline
Baseline & 39.2                        & 45.0                        & 36.0   & 36.9   & 39.3 \\
SD-AANet & {\color[HTML]{000000} 42.3} & {\color[HTML]{000000} 49.2} & 41.2   & 40.1   & 43.2 \\ \hline
\end{tabular}
\end{table}

\subsubsection{Ablation Study of Multi-class Segmentation}
\par To further explore the potential of SD-AANet, we propose a new pipeline for segmenting multi-class objects simultaneously under few-shot setting. We conduct experiments on Pascal-$5^i$ using baseline model and our SD-AANet. The experiments are based on 1-shot setting.

\par As show in Tab. \ref{tab7}, due to the difficulty of segmenting objects from multiply classes simultaneously, the results of multi-class 1-shot segmentation experiments are significantly lower than current few-shot segmentation results. However, Tab. \ref{tab7} still clearly shows the advantages of SD-AANet over the baseline model. On each fold, our SD-AANet can improve the performance of baseline model by at least 3.2 mIoU. For average result on all four folds, SD-AANet gets 3.9 mIoU increase compared to baseline.

\par We also analyze results of two models on each classes. As show in Tab. \ref{tab8}, "Class1" to "Class5" denote five classes in each folds orderly. We can see that for some classes, SD-AANet only gets slight improvement such as the "Class 5" of Fold-3, which is "tv/monitor". However, on some hard classes such as the "Class 4" of Fold-2, "motorbike", SD-AANet achieves remarkable progress.

\begin{table}[tbp]
\caption{Mutli-class 1-shot segmentation experiment results on each class.}
\label{tab8}
\centering
\resizebox{\linewidth}{!}{
\setlength{\tabcolsep}{1.4mm}{
\begin{tabular}{c|c|ccccc|c}
\hline
Fold index              & Method   & Class 1 & Class 2 & Class 3 & Class 4 & Class 5 & Mean \\ \hline
\multirow{2}{*}{Fold-0} & Baseline & 25.0    & 52.2    & 34.1    & 31.7    & 53.0    & 39.2 \\
                        & SD-AANet & 26.1    & 55.3    & 46.9    & 27.0    & 56.1    & 42.3 \\ \hline
\multirow{2}{*}{Fold-1} & Baseline & 38.4    & 56.9    & 15.1    & 58.0    & 56.6    & 45.0 \\
                        & SD-AANet & 42.0    & 61.7    & 16.3    & 64.1    & 61.9    & 49.2 \\ \hline
\multirow{2}{*}{Fold-2} & Baseline & 55.5    & 57.0    & 51.9    & 5.4     & 10.3    & 36.0 \\
                        & SD-AANet & 56.4    & 59.2    & 52.2    & 16.5    & 21.8    & 41.2 \\ \hline
\multirow{2}{*}{Fold-3} & Baseline & 57.9    & 35.9    & 52.3    & 19.8    & 18.9    & 36.9 \\
                        & SD-AANet & 61.5    & 40.7    & 57.3    & 21.8    & 19.1    & 40.1 \\ \hline
\end{tabular}}}
\end{table}

\subsection{Visualization Analysis}
\subsubsection{Qualitative Visualization of Segmentation Results}
To show the performance of our proposed architectures intuitively, we visualize final prediction masks produced by our SD-AANet in Fig. \ref{Fig7}. Meanwhile, we compare the segmentation results between baseline and SD-AANet to evaluate the performance improvement realized by SD-AANet. 
\par As shown in Fig. \ref{Fig7}, the columns (a), (f) are support images and their ground truths, which are marked in green in figures. The columns (b), (g) are query images and columns (c), (h) are the ground truths of them, which are also marked in green. The columns (d), (i) are the prediction results of baseline, and the columns (e), (j) are the predictions of SD-AANet, marked in red. 
\par As we can see, the second row in Fig. \ref{Fig7} shows cases which have tremendous differences between support objects and query target. Taking the (a) to (e) columns as an example, the bottles in support image and query image has totally different colors, shapes and perspectives, which leads to the segmentation failure of baseline. Relying upon the intrinsic feature extracted by SD-AANet, we can capture intrinsic features of the class and ignore the interference of other factors, so we successfully segment the bottle with negligible error. Other samples also confirm this point.

\subsubsection{t-SNE Visualization of Support Prototypes}
We conduct a t-distributed stochastic neighbor embedding (t-SNE) visualization experiment for support prototypes in Fig. \ref{Fig8}. In Fig. \ref{Fig8}, four columns in turn represent results of baseline, baseline with SAAM, baseline with SDPM and SD-AANet, and four rows in turn represent four folds from Fold-0 to Fold-3. We use models to process 5000 samples of 5 novel classes and get 5000 support prototypes output from SDPM or backbone. Then t-SNE is adopted to embed prototypes to 2-dimensional space to visualize, and the operation is repeated for 4 methods on 4 folds. As shown in Fig. \ref{Fig8}, SDPM can significantly expand the distance between prototypes of different classes and make prototypes of same classes more compact, so the results in columns (c) and (d) are more distinct. The columns (b) and (d) show SAAM can also make support prototypes more discriminative to a certain extent.
\par Taking Fold-1 and Fold-3 as examples, figures about two folds in first column show prototypes of 5 classes mix together and some classes are split at both ends of figures. The second column, which represents baseline with SAAM can distinguish classes slightly clearer, such as grey points in Fold-1 and orange points in Fold-3. In the third column, baseline with SDPM has greater performance to produce intrinsic prototype, so the points with same color seemed more compact such as grey points in Fold-1 and blue points in Fold-3. Combine the SAAM and SDPM, SD-AANet achieve the best performance which can obviously seen in figures. Clear dividing lines can be seen in the fourth column figures of Fold-1 and Fold-3.

\begin{figure}[htbp]
    \centering
    \includegraphics[width=0.9\linewidth]{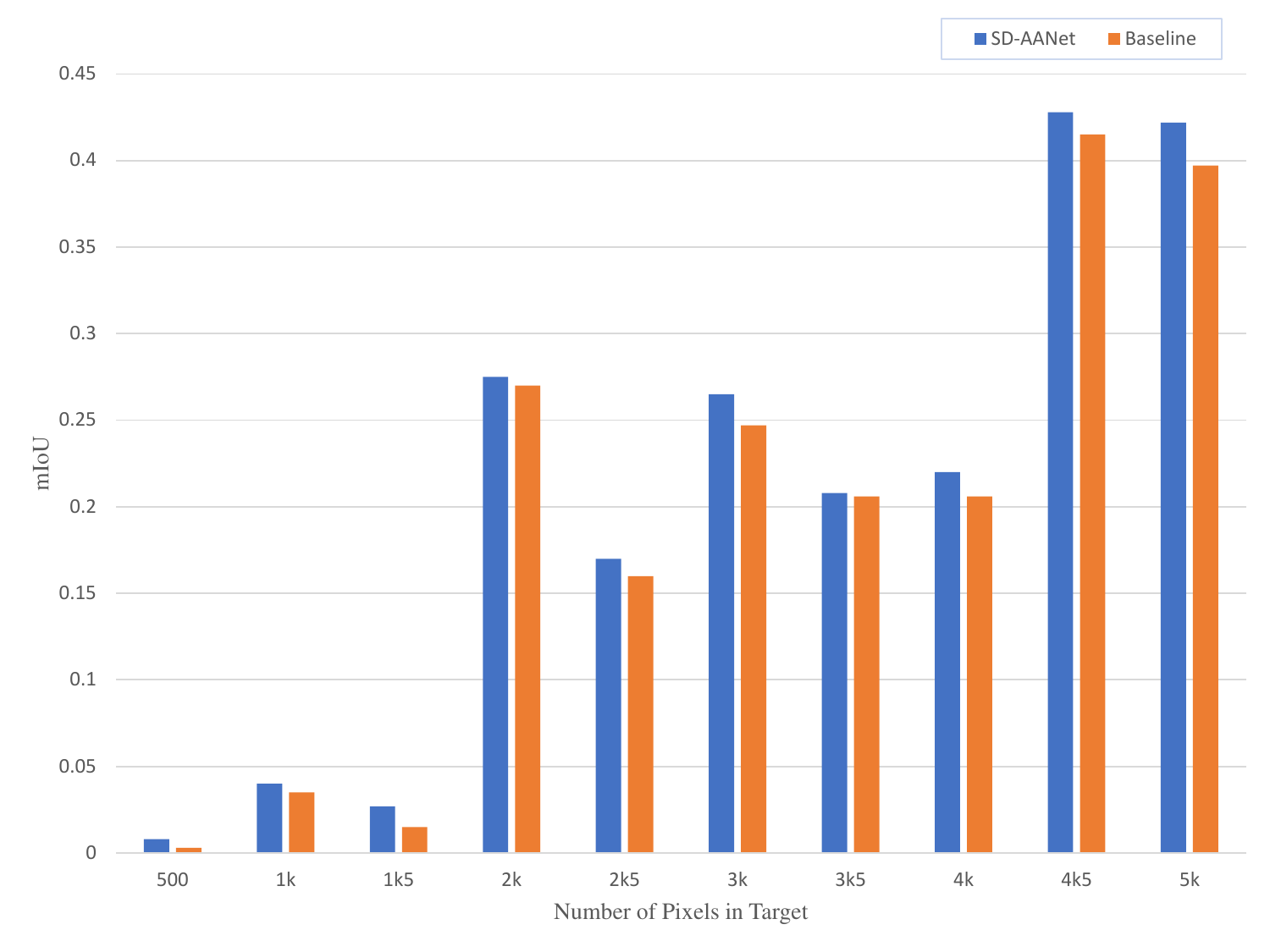}
    \caption{Quantitative comparative experiment of segmentation performance on small-scale target between baseline and SD-AANet on PASCAL-$5^i$ for 1-shot segmentation.}
    \label{Fig9}
\end{figure}

\subsubsection{Visualization of Performance on Small Target}
To further analyze the segmentation performance of SD-AANet for small targets, we choose the samples whose targets have less than 5000 pixels to conduct comparison experiment between SD-AANet and baseline. The samples are split to 10 parts, each part has a span of 500 pixels. We calculate the average class mIoU of each part produced by two models to draw a histogram, shown in Fig. \ref{Fig9}.
\par The results shown in Fig. \ref{Fig9} illustrate that our SD-AANet achieves more class mIoU in all 10 parts, so SD-AANet has greater segmentation performance than baseline for small target segmentation.

\subsubsection{Visualization of Affinity Attention}
\begin{figure}[htbp]
    \centering
    \includegraphics[width=0.9\linewidth]{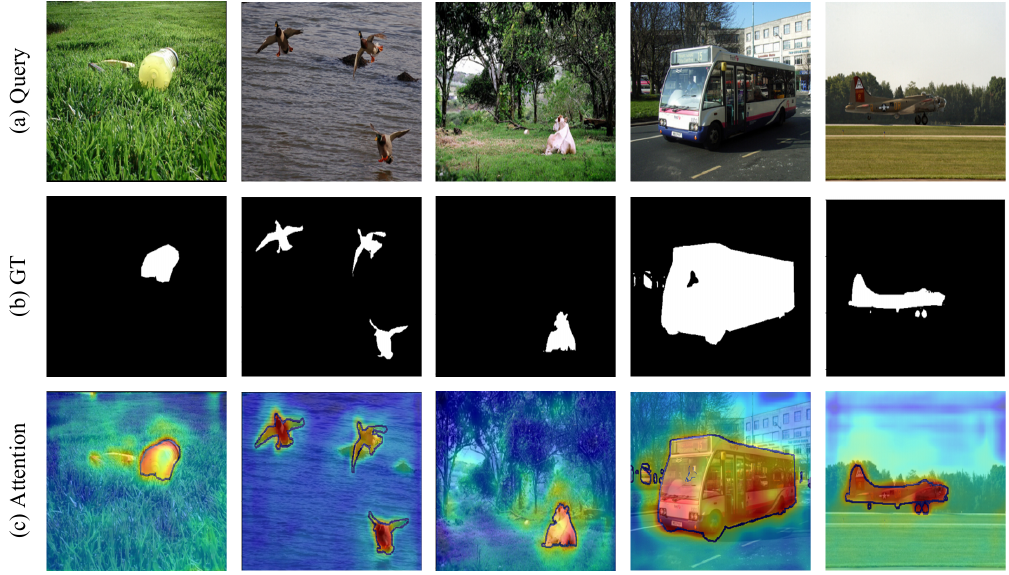}
    \caption{Visualization of affinity attention map generated by SAAM on PASCAL-$5^i$ for 1-shot segmentation.}
    \label{Fig10}
\end{figure}

To intuitively analyze the quality of affinity attention produced by SAAM, we visualize the attentsion map as shown in Fig. \ref{Fig10}. The first two rows are query image and its ground truth label respectively, and the third row is affinity attention map where close to red (warm-toned) means more attention, vice versa.
\par In first three columns, we can see that SAAM can effectively capture the spatial information of targets, even if they are small or there are multiple targets in one image. The last two columns show the SAAM can focus on large-scale targets and capture key information of separate parts, such as the rear-view mirror of the bus and wheels of the aeroplane, with only one support sample.

\subsubsection{Visualization of Prototype's Representative}
\par To discuss the representative of support prototype, we calculate the cosine similarity between support prototype and its feature map
\begin{equation}
    cos\left ( x_{i, j}, p_{s}^{'} \right ) = \frac{x_{i, j}^{T}\cdot p_{s}^{'}}{\left \| x_{i, j} \right \|\cdot \left \|  p_{s}^{'}\right \|}
\label{eq13}
\end{equation}
where $x_{i, j}$ denotes the vector in support feature at $\left( i, j\right)$ location and $x_{i, j}^{T}$ denotes its transpose. $p_{s}^{'}$ denotes the support prototype output from SDPM. Finally the $\left \| \cdot \right \|$ denotes the norm of the vector.
\par
\begin{figure}[htbp]
    \centering
    \includegraphics[width=0.9\linewidth]{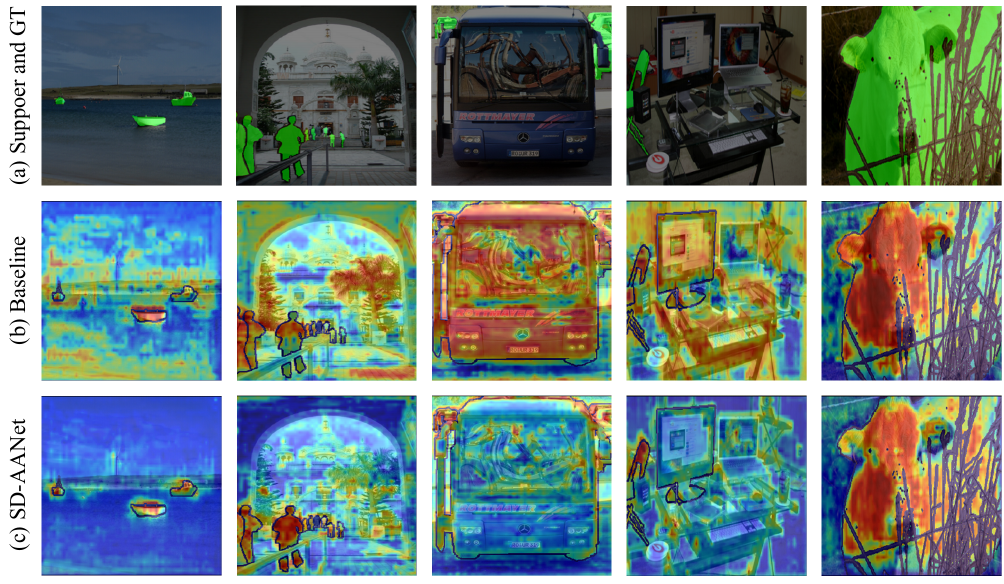}
    \caption{Visualization of support similarity map generated by support prototype and support feature using cosine similarity. The comparison visualization study is process on PASCAL-$5^i$ for 1-shot segmentation.}
    \label{Fig11}
\end{figure}

 Similar to attention maps in Fig. \ref{Fig10}, the similarity map shown in Fig. \ref{Fig11} use warm-toned color to represent high similarity.  Three rows from top to bottom in turn illustrate support image and its ground truth, similarity map produced by baseline and similarity map produced by SA-AANet. In first four columns we can see the prototype produced by SD-AANet can filter more irrelevant background compared to baseline, which means the prototype focus on intrinsic feature to capture the target regardless of other environmental factors. In fifth column, baseline fails to found part of sheep which is mixed up with the fence, while SD-AANet finds the whole spatial area of the sheep.

\subsubsection{Visualization of Failure Cases}
\begin{figure}[htbp]
    \centering
    \includegraphics[width=0.9\linewidth]{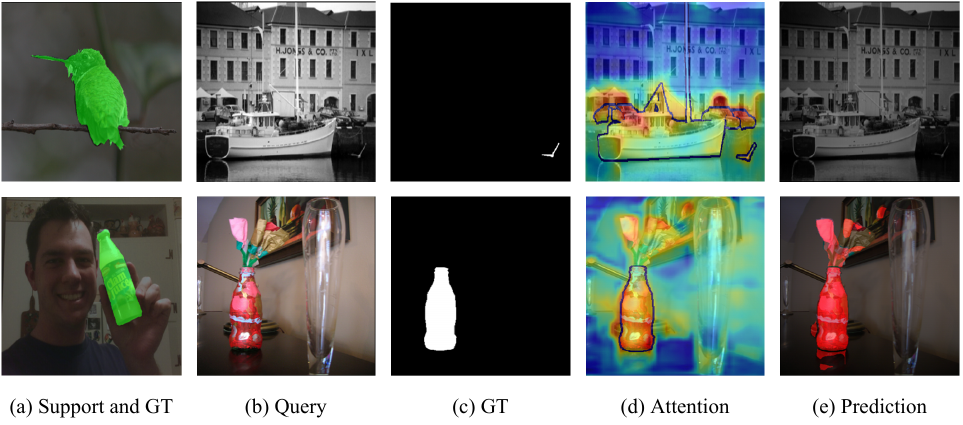}
    \caption{Visualization study of failure cases of SD-AANet on PASCAL-$5^i$ for 1-shot segmentation.}
    \label{Fig12}
\end{figure}

As shown in Fig. \ref{Fig12}, SD-AANet still fails to segment some targets because the target size is too small or the target is very similar to the background. The first column is support image and its ground truth, and the next two columns are query image and its ground truth respectively. The fourth column is attention map produced by SAAM and the last column is prediction of SD-AANet. 

\par We can see in the first row that the target is a bird of very small size, and there are great differences between support object and query target. In second row, the bottle is very similar to the flower which is hard to discriminate.

\section{Conclusion}
In this paper, we propose a novel few-shot segmentation method named SD-AANet. Our method significantly differs from existing methods by combining self-distillation, prototypical learning and affinity learning. To address the problem of large intra-class variation, we propose the self-distillation guided prototype module (SDPM) to extract intrinsic prototype, which can efficiently align the features of support and query. We further construct a supervised affinity attention module (SAAM) for generating high quality prior attention map of query image. Extensive experiments on two standard benchmarks verify the performance superiority of our method. Besides, future work may focus on utilizing self-distillation to zero-shot segmentation task.

\bibliographystyle{ieee}
\bibliography{manuscript}

\begin{IEEEbiography}[{\includegraphics[width=1in,height=1.25in,clip,keepaspectratio]{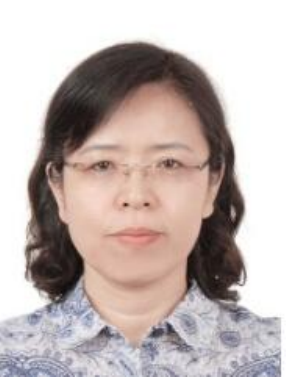}}]{Qi Zhao}
was born in China, in 1966. She received Ph.D degree in communication and information system from Beihang University, Beijing, China. 
\par She is a professor and works in Beihang University. She was in the Department of Electrical and Computer Engineering at the University of Pittsburgh as a visiting scholar from 2014 to 2015. Since 2016, she has been working on wearable device based first-view image processing and deep learning based image recognition. Her current research interests include few-shot semantic segmentation, medical image processing, object detection and target tracking.
\end{IEEEbiography}
\vspace{-10 mm}
\begin{IEEEbiography}[{\includegraphics[width=1in,height=1.25in,clip,keepaspectratio]{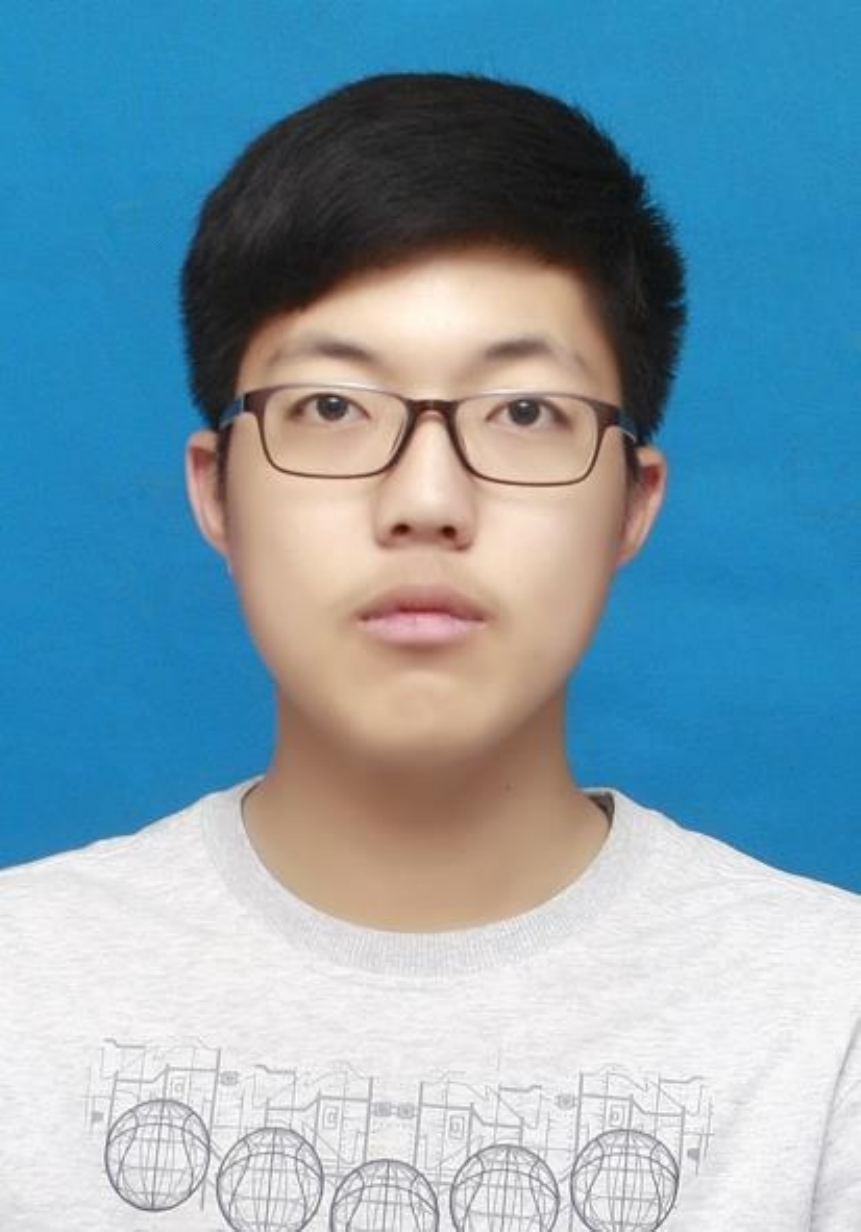}}]{Binghao Liu}
received B.E. degree in electronics and information engineering from Beihang University, Beijing, China, in 2019. He is currently pursuing the Ph.D. degree with the School of Electronic and Information Engineering, Beihang University, Beijing. His research interests include weakly supervised learning, semantic segmentation and few-shot segmentation.
\end{IEEEbiography}
\vspace{-10 mm}

\begin{IEEEbiography}[{\includegraphics[width=1in,height=1.25in,clip,keepaspectratio]{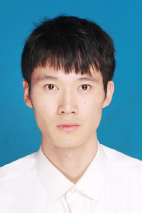}}]{Shuchang Lyu}
received the B.E. degree in communication and information from Shanghai University, Shanghai, China, in 2016, and the M.E. degree in communication and information system from the School of Electronic and Information Engineering, Beihang University, Beijing, China, in 2019. He is currently pursuing the Ph.D. degree with the School of Electronic and Information Engineering, Beihang University, Beijing. His research interests include deep learning, image classification, few-shot semantic segmentation and object detection.
\end{IEEEbiography}
\vspace{-10 mm}
\begin{IEEEbiography}[{\includegraphics[width=1in,height=1.25in,clip,keepaspectratio]{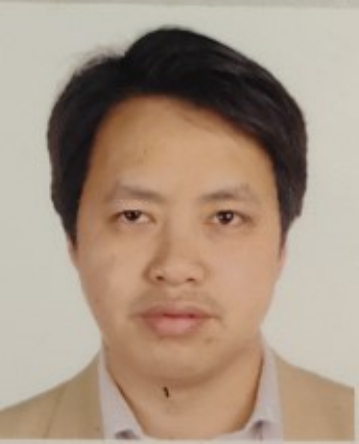}}]{Huojin Chen}
was born in China, in 1967. He received the bachelor's degree in Testing Technology and Instruments from Jilin University of Technology, Changchun, China, in 1989, and the master's degree in System Engineering from Tianjin University, Tianjin, China, in 1992. 
\par He is a Associate Professor with the College of Art and Design, Beijing University of Technology. His main research interests focus on Computer Vision, System Engineering and Design Management.

\end{IEEEbiography}

\end{document}